\documentclass[runningheads]{llncs}
\usepackage[T1]{fontenc}
\usepackage{graphicx}
\usepackage{subcaption}
\usepackage{amssymb}
\usepackage{amstext}
\usepackage{amsmath}
\usepackage{multirow}
\usepackage{times} 
\usepackage{booktabs}
\usepackage{makecell}

\usepackage[numbers,sort&compress]{natbib}

\usepackage{algorithm}
\usepackage[hidelinks]{hyperref}
\usepackage{algorithmic}
\usepackage[textsize=tiny]{todonotes}
\setuptodonotes{inline} 
\usepackage[bottom]{footmisc}
\usepackage{url,cleveref} 

\newcommand{\mll}{<\!\!<} 

\begin{document}
\title{Towards Scalable Lottery Ticket Networks using Genetic Algorithms\protect\footnotetext{An earlier version of this work was presented at the International Conference on Neural Computation Theory and Applications (NCTA 2024) \cite{ncta24}. This article extends our conference paper with updates to the method to improve multi-class classification, an expanded experimental setup, and a multi-class performance anaylsis with visual and analytical justifications.}}

\titlerunning{Towards Scalable Lottery Ticket Networks using Genetic Algorithms}
%
\author{Julian Schönberger \and Maximilian Zorn \and  Jonas Nüßlein \and Thomas Gabor \and  \\ Philipp Altmann}
\authorrunning{J. Schönberger et al.}
%
\institute{LMU Munich, Munich, Germany\\
\email{julian.schoenberger@ifi.lmu.de}}
\maketitle              
\begin{abstract}
Building modern deep learning systems that are not just effective but also efficient requires rethinking established paradigms for model training and neural architecture design. Instead of adapting highly overparameterized networks and subsequently applying model compression techniques to reduce resource consumption, a new class of high-performing networks skips the need for expensive parameter updates, while requiring only a fraction of parameters, making them highly scalable. The Strong Lottery Ticket Hypothesis posits that within randomly initialized, sufficiently overparameterized neural networks, there exist subnetworks that can match the accuracy of the trained original model—without any training. This work explores the usage of genetic algorithms for identifying these strong lottery ticket subnetworks. We find that for instances of binary and multi-class classification tasks, our approach achieves better accuracies and sparsity levels than the current state-of-the-art without requiring any gradient information. In addition, we provide justification for the need for appropriate evaluation metrics when scaling to more complex network architectures and learning tasks.

\end{abstract}
%
%
%

\newpage
\section{Introduction}\label{sec:intro}
\begin{figure}[b]
  \centering
  \includegraphics[width=0.9\linewidth]{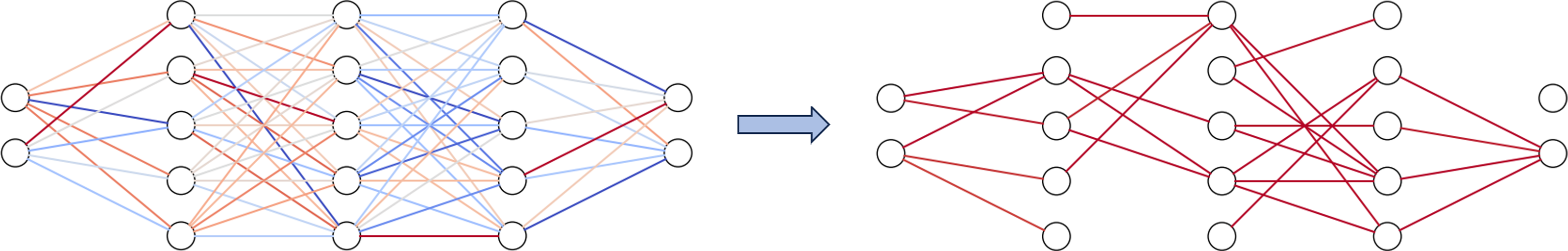}
  \caption{Visualization of a lottery ticket network \cite{ncta24}. \textbf{Left}: Full network graph, where red connections are those that consistently appear in most evolved lottery ticket networks within a sample population, while blue connections do not. \textbf{Right}: An example of an evolved lottery ticket subnetwork, retaining only a subset of active connections.}
  \label{fig:lottery_ticket}
\end{figure}
Modern deep learning models consist of billions of parameters, that require up to millions of optimization steps to train, taking weeks or even months on high-performance computing hardware. The consequences include, but are not limited to, high economic expenses and excessive energy consumption, often leading to a substantial amount of $CO_2$ emissions. Future Deep Learning models need to be more efficient and effective to be sustainable in the long-term. 
\\[2pt]
We argue that the massive overparameterization of today's models, which is one major contributor to the success of deep learning is also a significant source of inefficiency, but may inadvertently provide a path to a new class of highly scalable model architectures. This overparameterization allows for the existence of \textit{strong lottery tickets} — subnetworks within the randomly initialized larger model that achieve comparable performance to the full network without any training \citep{ramanujan2020s}, an example of which is illustrated in Fig.~\ref{fig:lottery_ticket}. By simply pruning parameters (i.e., setting their values to zero), we can shift the model to an optimal region in the parameter space, without having to perform small-scale updates to the entire network parameterization, thereby reducing the computational cost and potentially improving efficiency. This alternative paradigm to classical training not only has the potential to reduce the number of optimization steps required to achieve a comparable or better performance, but also leads to a reduction in model complexity (w.r.t. the model size). 
\\[2pt]
The efficacy of this approach can be further enhanced by using a sophisticated optimization algorithm that fully leverages the new discrete nature of the optimization problem. In this work, we argue for the benefit of using \textit{genetic algorithms} (GAs) as capable combinatorial optimization methods that can explore large areas of the complex solution space. In contrast to gradient-based methods, genetic algorithms are architecture-agnostic, meaning they can be applied to any \textit{artificial neural network} (ANN) architecture regardless of its structure or layer types. This flexibility enables searching for subnetworks in a diverse range of models, from simple feedforward networks to large scale transformer architectures. Furthermore, since these algorithms do not rely on gradient information, they can be applied to non-differentiable models or novel architectures where gradient-based methods might not be directly applicable.
\newpage
\noindent This work presents an extensive analysis of the significance of using genetic algorithms for identifying strong lottery tickets in randomly initialized feed-forward neural networks. It forms the basis for future research on more complex model architectures. After explaining the major components of our algorithm, we perform a comprehensive performance evaluation on two binary as well as two multi-class classification tasks. We complement our findings with the comparative performance of the  state-of-the-art as well as networks obtained through standard training. We extend a previous version of this paper (cf. \cite{ncta24}) by providing empirical and analytical justifications for the use of a different performance criterion in solving the multi-class classification tasks, as well as detailed evaluations of said tasks. In addition, we incorporate a ``post-evolutionary pruning'' routine that further improves the accuracy-sparsity trade-off present during optimization. 
\\
Our main contributions include the successful application of a genetic algorithm to identify strong lottery ticket networks for binary and multi-class classification tasks, surpassing state-of-the-art accuracies and sparsities in several instances. Additionally, we provide both visual and analytical insights into the inherent properties of strong lottery tickets. 

\section{Related Work}\label{sec:rel_work}
The Lottery Ticket Hypothesis has attracted significant attention in recent years, leading to the identification of various connections to related fields. In this section, we will explore the existing literature and discuss its relevance to our work.

\paragraph{\textbf{Lottery Ticket Hypothesis}}\label{sec:rel_work:lth}
Frankle and Carbin~\cite{frankle2018lottery} discovered that a network, after being pruned and having its remaining weights reset to their original random initialization, could be retrained to achieve test accuracy comparable to the original network in a similar number of training iterations. They termed this phenomenon the \textit{Lottery Ticket Hypothesis} (LTH) and referred to the pruned subnetwork as a \textit{winning ticket}. They proposed an algorithm based on iterative magnitude pruning to identify these winning tickets. Since then, numerous methods have been developed to identify these subnetworks. For instance, Jackson et al. \citep{jackson2023finding} employed an evolutionary algorithm where fitness is calculated based on network density and validation loss, addressing the trade-off between subnetwork sparsity and accuracy. Other subsequent studies \citep{zhou2019deconstructing, wang2020pruning} extended the LTH, demonstrating that it is possible to find subnetworks within randomly initialized networks that perform better than random guessing, even without training. Zhou et al. \citep{zhou2019deconstructing} introduced neural network masking as an alternative training method and introduced the concept of ``supermasks.''
\paragraph{\textbf{Strong Lottery Ticket Hypothesis}}
Ramanujan et al.~\citep{ramanujan2020s} expanded on this concept by introducing the Strong Lottery Ticket Hypothesis (SLTH), which posits that a sufficiently overparameterized neural network with random initialization contains a subnetwork, called \textit{strong lottery ticket} (SLT), that can achieve competitive accuracy (relative to the large, trained network) without requiring any training \citep{malach2020proving}. In addition, they proposed the \emph{edge-popup} algorithm as a method for identifying strong lottery tickets. It utilizes a gradient estimation technique to approximate the gradient of a pop-up score for each network weight. These pop-up scores are updated through stochastic gradient descent (SGD). A series of theoretical works explored the necessary degree of overparameterization \citep{malach2020proving, orseau2020logarithmic, pensia2020optimal}, demonstrating that logarithmic overparameterization is sufficient \citep{orseau2020logarithmic, pensia2020optimal}. In their pursuit of more efficient methods for discovering SLTs, Whitaker \citep{whitaker2022quantum} proposed three theoretical quantum algorithms based on edge-popup, knowledge distillation \citep{hinton2015distilling}, and NK Echo State Networks \citep{whitley2015optimal}.
Finally, Chen et al. \citep{chen2021peek} introduced a new category of high-performing subnetworks called ``disguised subnetworks''. Unlike regular SLTs, these subnetworks must first be ``unmasked'' through specific weight transformations. The authors identify these unique subnetworks using a two-step algorithm that applies sign flips to the weights of pruned networks, leveraging Synflow \citep{tanaka2020pruning}. By leveraging these additional weight transformations, they obtain subnetworks that have greater capacity than regular SLTs.
\paragraph{\textbf{Weak Lottery Ticket Hypothesis}}
A limited number of methods for discovering strong lottery tickets have been developed so far, with most empirical research focusing on the original Lottery Ticket Hypothesis. These methods identify so-called \textit{weak lottery tickets}, which can achieve competitive accuracies on much smaller subnetworks, but only after re-training the subnetworks' weights. This training-pruning-retraining cycle is often resource-intensive, and the advantages over standard training are not always clear. Conversely, the search for strong lottery tickets enables the discovery of high-accuracy subnetworks without the need for expensive retraining steps. Moreover, when integrated with meta-heuristic optimization, this methodology can be employed in the context of structures exhibiting discontinuous functions, a domain in which gradient-based methods encounter significant challenges. 
In this paper, we propose a novel approach for identifying strong lottery tickets, based on genetic algorithms, leveraging the principles of biological evolution. In contrast to conventional methods that often employ heuristics and pseudo-training algorithms with gradient descent and pre-defined pruning rates, our approach does not necessitate gradient information. It enables direct optimization of the subnetwork structure, without imposing artificial constraints on the maximum number of pruned weights. 
\\[2pt]
Genetic algorithms have been shown to be highly effective in the resolution of NP-hard problems and are particularly well-suited for the optimization of non-convex objective functions with numerous local minima, saddle points, and plateaus. The optimization landscape of the Strong Lottery Ticket Hypothesis is complex, influenced by factors such as masking, random initialization, the scale of the feature space and the objective function.
It is worth noting that Jackson et al. \citep{jackson2023finding}, while using an approach similar to ours, apply their evolutionary algorithm to the original LTH and thus can only identify weak lottery tickets that require retraining.

\paragraph{\textbf{Extreme Learning Machine}}
Huang et al. \citep{huang2006extreme} introduced the Extreme Learning Machine (ELM), a method conceptually similar to the Strong Lottery Ticket Hypothesis, where the random parameter values of the hidden layer in a single-hidden-layer neural network are fixed. The optimal weights for the output layer are then computed using the closed-form solution for linear regression.
In contrast to SLTs, dense models such as ELM are less parameter-efficient, face challenges in scaling to deeper architectures (which often require complex adaptations, such as those based on autoencoders \citep{kasun2013representational}), and involve the computation of matrix inverses, which is computationally expensive.

\paragraph{\textbf{Neural Architecture Search}}
The process of neural architecture search (NAS) shares certain similarities with the search for lottery tickets, given that both involve the generation of networks with previously unknown structures and untrained (though potentially selected) weights. Gaier and Ha \citep{gaier2019weight} examined the impact of network architecture versus parameter initialization on task performance. Upon initializing all parameters with a single value drawn from a uniform distribution, they discovered that certain architectures attain higher accuracy than random on the MNIST dataset. In their seminal work, Wortsman et al. \citep{wortsman2019discovering} proposed a method that facilitates the continuous adaptation of a network's connection graph and its associated parameters during the training process. Their innovative approach demonstrated that the resulting networks exhibited superior performance in comparison to both manually designed and randomly structured networks. In contrast to our approach, Gaier and Ha \citep{gaier2019weight} used a single fixed value for the parameters, rather than sampling from a random distribution. The method introduced by Wortsman et al. \citep{wortsman2019discovering} offers an alternative to finding winning tickets. Ramanujan et al. \citep{ramanujan2020s} later introduced edge-popup, inspired by Wortsman et al.'s work, but since their approach involves learning the network structure and its parameterization together, it cannot be applied for finding pruning masks for strong lottery tickets.
%
\paragraph{Evolutionary Pruning}
Unlike NAS or neuroevolution, which typically involve evolving the network's topology, evolutionary pruning focuses solely on removing connections and possibly entire neurons from the network graph. With these techniques, networks can often be reduced in size without sacrificing performance. This area of research includes methods that differ in their choice of solution representation (direct or indirect encoding) and the number of objectives considered. Direct encoding methods frequently utilize binary masks that are applied to network structures, such as individual weights or convolution filters \citep{wu2021differential}. Common multi-objective tasks include not only sparsity but also accuracy improvement or energy efficiency \citep{wang2021evolutionary}. Our approach employs binary pruning masks with two objectives in mind: accuracy and sparsity. To the best of our knowledge, we are the first to apply evolutionary pruning within the context of the SLTH.

\paragraph{Other Pruning Methods}
As noted by Wang et al. \citep{wang2021recent}, in addition to the classic LTH, which uses static pruning masks on trained networks, and the SLTH, which bypasses training entirely, there exists a third category of methods that prune networks at initialization using pre-selected masks \citep{lee2018snip, wang2020picking, tanaka2020pruning}. For instance,  Lee et al. \citep{lee2018snip} proposed a pruning mask created before training, which removed structurally insignificant connections based on a new saliency criterion known as connection sensitivity. Similar to our approach, their method is one-shot, as the network only requires a single pruning step; however, training is still involved, and specific pruning criteria are needed to identify optimal subnetworks.

\section{Method}
\label{sec:method}
The components of the genetic algorithm are examined in detail below. This examination includes the configuration of candidate solutions, the evaluation process, the selection of parents and survivors and the series of genetic operations that govern the evolutionary process.

\paragraph{Solution Representation}
The generation of strong lottery ticket networks is achieved by implementing an evolutionary algorithm, assuming that the intended task of the network is predetermined, (e.g., specified by an objective function such as a classification accuracy function or a loss function, denoted by $\mathcal{F}$). The architecture graph of the complete network and the vector of its randomly initialized parameters, $\mathbf{w} = \langle w_0, ..., w_n \rangle$ with $w_i \in \mathbb{R}$ for all $i$, are also provided. The proposed approach then generates a bit mask (genotype), denoted by $\mathbf{b} = \langle b_1, ..., b_n \rangle$ with $b_i \in \{0, 1\}$ for all $i$. This mask is used to construct a masked network (subnetwork), or phenotype, represented by $\mathbf{w'} = \langle b_i \cdot w_i \rangle_{i=1,...,n}$. These subnetworks are typically considerably smaller than the original networks w.r.t. non-zero weights, without significant performance degradation w.r.t. $\mathcal{F}$. In formal terms, let $\mathbf{w^*}$ denote the $n$ weights of the trained full network, then it follows that $\mathbf{b}$ should satisfy the following equation: $\sum_{i=0}^n b_i \mll n$ and $\mathcal{F}(\mathbf{w'}) \approx \mathcal{F}(\mathbf{w^*})$. It is important to note that the following discourse refers exclusively to the weights within the parameter vector. It does not encompass any potential bias nodes. Even though they are not pruned, biases are still initialized the same way as the weights.

\paragraph{Fitness and Selection}
In order to promote the development of strong lottery tickets, a lexical evolutionary optimization process is employed. Two objectives are pursued: We primarily focus on identifying subnetworks that attain the level of accuracy accomplished through conventional training; subsequently, we also promote the identification of subnetworks that are as sparse as possible without compromising accuracy. This multi-objective approach enables the pruning of subnetworks to a considerable extent, even in scenarios where high accuracies have already been attained. An intriguing characteristic of using an evolutionary optimization approach compared to gradient-based optimization is that we can directly optimize the accuracy function without having to rely on the loss function intermediary (i.e. $\mathcal{F} = \mathcal{A}$). Our later findings prove that this approach can be very effective for binary classification problems, but appears to reach its limits when the number of classes increases (cf. \ref{subsec:prob_multiclass}). Fortunately, our primary performance metric can be exchanged seamlessly, such that contrary to maximizing the accuracy function we can e.g. minimize a loss function instead (i.e. $\mathcal{F} = \mathcal{L}$). The assessment of individuals within the evolutionary framework takes place in two processing steps: In the first step, the objective is to identify parents (i.e., individuals suitable for recombination). In this stage, we solely focus on the performance goal. Conversely, in the second step, the objective is to select survivors (i.e., individuals suitable for the next generation). In this stage, we also consider the sparsity objective. This differentiation reflects the observation that recombination is the main contributor for producing higher-performing individuals over the course of the evolution. Focusing on the performance goal (i.e. accuracy/loss) in parent selection leads to an effective prioritization strategy: the fitness of individuals is determined by the measured performance on the training dataset, and individuals are then ranked accordingly. Although sparsity is considered in survivor selection, performance remains the primary factor: Within groups of individuals with the same performance level, those with higher sparsity are preferred.
\\[2pt]

\noindent We employ an (elitist) cut-off selection method for the purpose of survivor selection. \footnote{Alternative selection methods, including roulette and random walk selection, were also experimented with; however, it was found that the selection method employed did not have a significant impact.} This method entails the selection of the top $k$ individuals from the current population and their subsequent transfer to the following generation's population. For our GA it holds that $k = N$ where $N$ denotes the original population size. 

\noindent It was observed that, due to the fact that our defined genetic operators don't operate ``in place'', the population frequently surpassed its initial size $N$ during generation transitions, which necessitated the reduction of the population size for the subsequent generation. Concerning the process of parent selection, any individual has the capacity to be selected as a first parent, with probability $\texttt{rec\_rate} \in [0,1]$, and subsequently paired with a second parent, selected at random from the top $l$ individuals in the present population, where $l = N \cdot \texttt{par\_rate}$ where $\texttt{par\_rate}$ is another hyperparameter.

\vspace{-2pt} 
\paragraph{Genetic Operators}\label{subsec:gen_and_var}
The initial population is generated in two steps. First, individuals are uniformly generated at random, meaning that each bit has an equal chance of being selected at any position in the binary pruning mask. Second, from the individuals that were randomly generated, those that don't reach a certain accuracy bound are discarded. In our implementation, an adaptive bound undergoes dynamic reduction if an insufficient number of individuals fulfill the specified boundary value in a given time frame. This process follows the outline of a predefined exponential function. This is done to lessen the impact of random sampling on runtime. The employment of an adaptive accuracy bound enables the establishment of an initial bound that is higher compared to what is tolerable with a static bound, and it has been demonstrated to result in superior final accuracies. In the subsequent discussion, the configuration that executes solely the initial step we will refer to as \textbf{GA}, and the configuration where an adaptive accuracy boundary is used (i.e. the first and second step) we will call \textbf{GA (adaptive AB)}. \footnote{\textbf{Code} for the GA and our experiments is available at \url{https://github.com/julianscher/SLTN-GA}.} \\[2pt]
We implement a single-point mutation process, wherein individuals are randomly selected from the current population at a rate denoted by $\texttt{mut\_rate}$ and a mutant is generated through a random bit flip. For recombination, we employ a random crossover mechanism between two parents. It is noteworthy that mutants and offspring are directly incorporated into the population, with no direct replacement of their source individuals. In each generation, we introduce $m$ newly generated individuals to the population to further enhance diversity within the population. The value of $m = N \cdot \texttt{mig\_rate}$ is defined by the hyperparameter $\texttt{mig\_rate}$.

\newpage
\paragraph{Post-Evolutionary Pruning}\label{subsec:post_evol_prune}
Since our fitness evaluation and selection criteria prioritize the performance objective, we observe that after the GA has terminated, the possibilities for pruning some of the remaining connections are typically not yet fully exhausted. That's why we employ an additional simple post-processing routine that sequentially runs through the final bit mask, setting all those bits to zero that do not exert a negative influence on the accuracy. Depending on factors like the initial sparsity, the prune-rate used to create the individuals for the initial population and the duration of the evolution, the post-processing routine has demonstrated to enhance the sparsity of the final individuals by a margin of multiple percentage points and can even lead to minor additional accuracy improvements (cf. Fig.~\ref{fig:sparsities_digits}).


\section{Experimental Setup}\label{sec:exp}
To evaluate the performance of the above genetic algorithm in finding SLTs, we apply it to several datasets of varying complexity and different network architectures with an increasing number of parameters. In addition to binary classification problems, we also include two multi-class classification tasks. To complement our GA results, we compare the performance with the state of the art and with networks trained with standard backpropagation.

\paragraph{Hyperparameters}
In the subsequent series of experiments, the genetic algorithm is implemented with a fixed population size of $N = 100$ individuals, while the rates for parent selection, recombination, mutation, and migration remain constant. For the process of recombination, a value of 0.3 is employed, which dictates that approximately $30\%$ of the total population is selected to become a first parent. For the purpose of generating offspring, a recombination partner is randomly selected from the top 30\% of the population due to the recombination rate of $\texttt{rec\_rate} = 0.3$.  The mutation rate, set at $\texttt{mute\_rate} = 0.1$, ensures that approximately $10\%$ of individuals generate a mutant that is added to the population, which is a relatively high value. This is done with the intention of generating highly exploratory runs. The migration rate, defined as $\texttt{mig\_rate} = 0.1$, is set to ensure that approximately $10\%$ of the individuals in the interim population prior to survivor selection are newly generated. Table \ref{tab:GA_hyperparameters} provides a summary of the selected hyperparameter values. The termination of the GA requires the evolution of the population for a minimum of 100 generations and activates when there is no discernible improvement in performance for the past $50$ generations. When employing the GA (adaptive AB), we limit the evolution to a maximum of 200 generations, as we have observed that beyond this point, the improvement in performance is typically negligible. An explicit hyperparameter search was not conducted to determine optimal values; instead, decisions were based on observations made during the implementation phase. This approach deliberately focuses on the expected performance rather than the hypothetically maximal one.

\begin{table}[h]
    \centering
    \small
    \begin{tabular}{cc}
        \toprule
         Hyperparameter &  Value \\
         \midrule
         $\texttt{pop\_size}$ $N$ & 100 \\
         $\texttt{rec\_rate}$ & 0.3 \\
         $\texttt{par\_rate}$ & 0.3 \\
         $\texttt{mut\_rate}$ & 0.1 \\
         $\texttt{mig\_rate}$ & 0.1 \\
         \bottomrule
         \\
    \end{tabular}
    \caption{The hyperparameters utilized for our GA evaluation. Table taken from \cite{ncta24}.}
    \label{tab:GA_hyperparameters}
\end{table}

\begin{figure*}\centering
  \subfloat[Moons dataset\label{fig:moons}]{\includegraphics[width=0.46\linewidth]{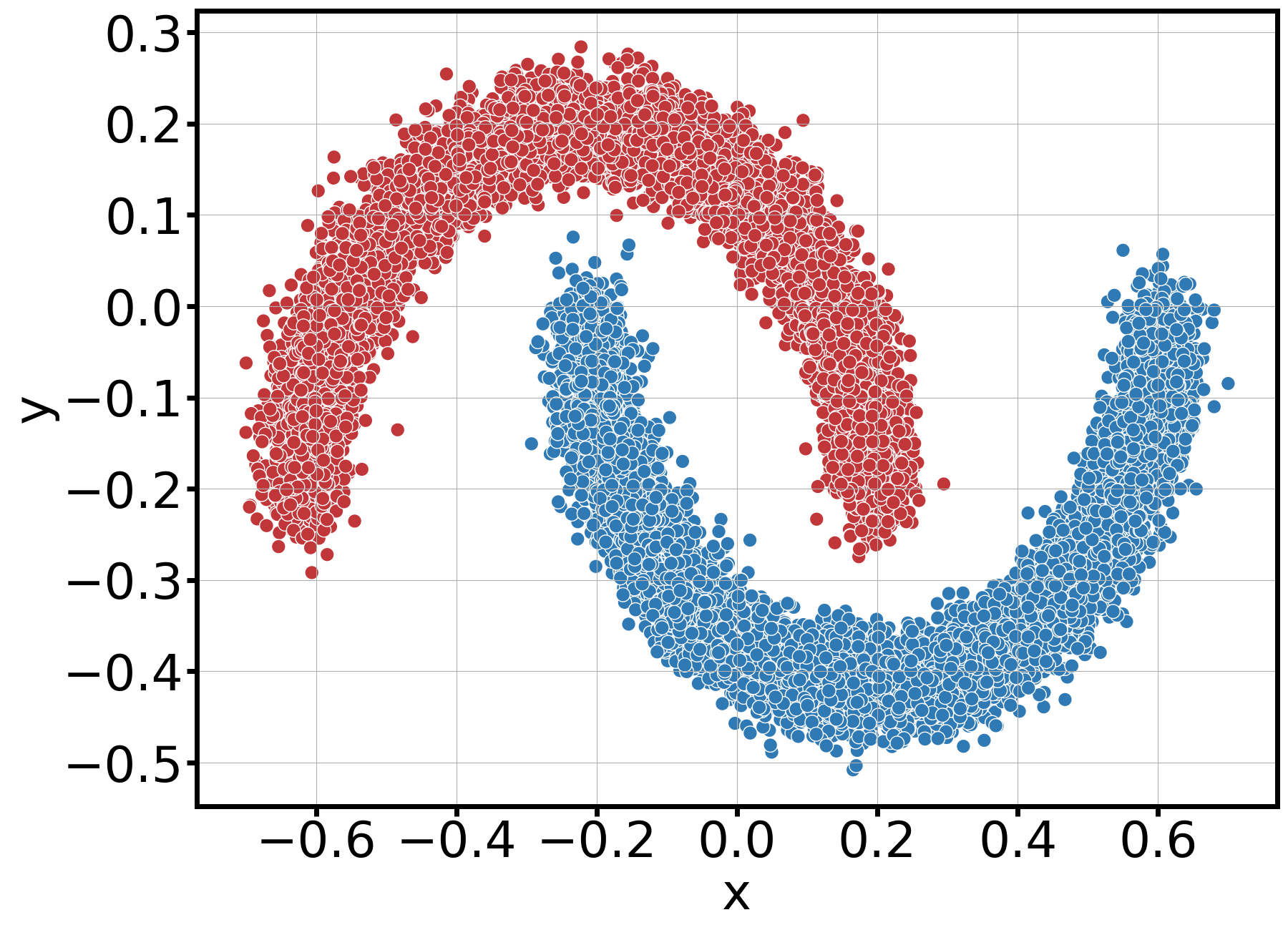}}\hspace{16pt}
  \subfloat[Circles dataset\label{fig:circles}]{\includegraphics[width=0.46\linewidth]{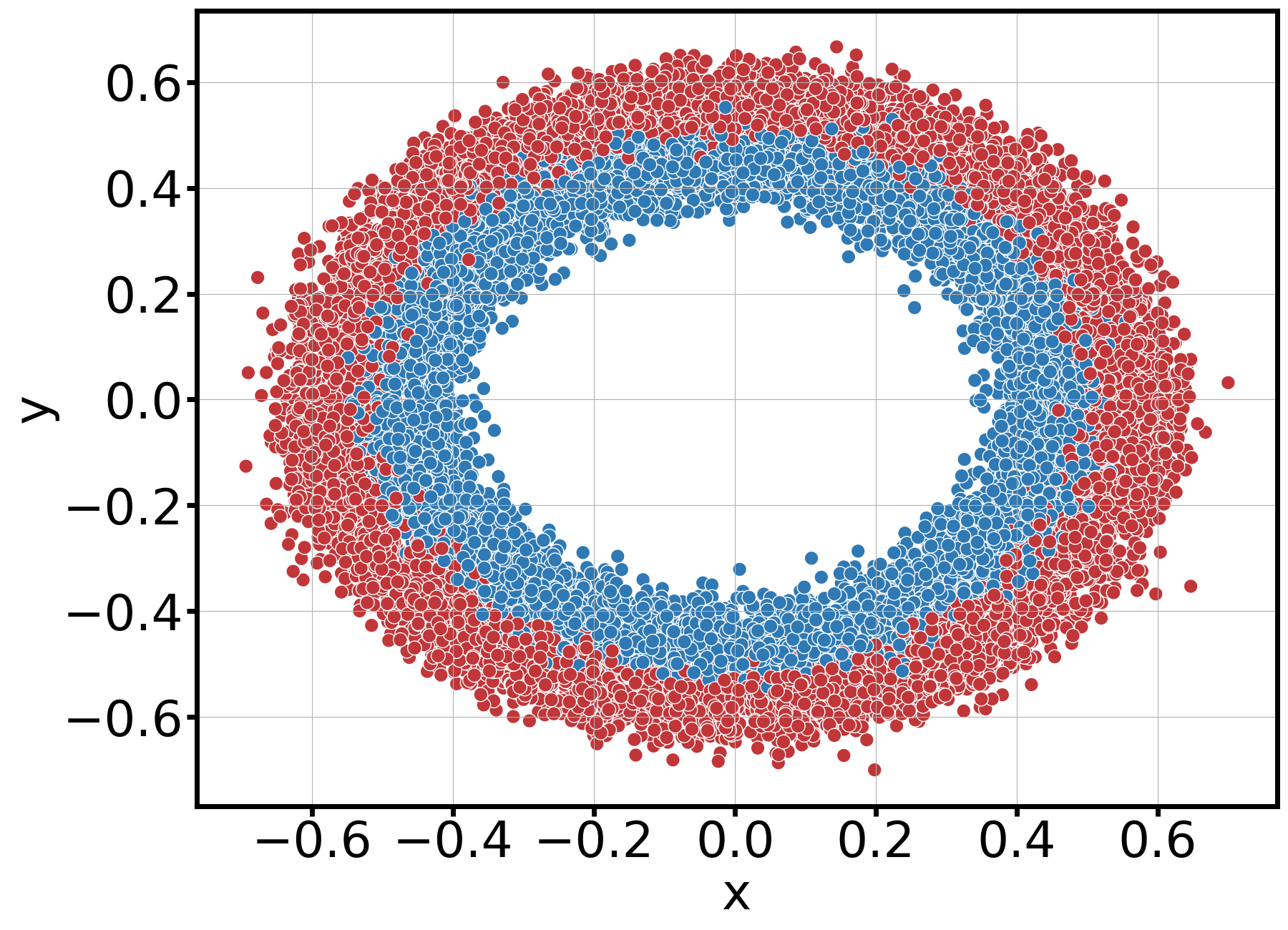}}\\[8pt]
  \subfloat[Blobs dataset\label{fig:blobs}]{\includegraphics[width=0.46\linewidth]{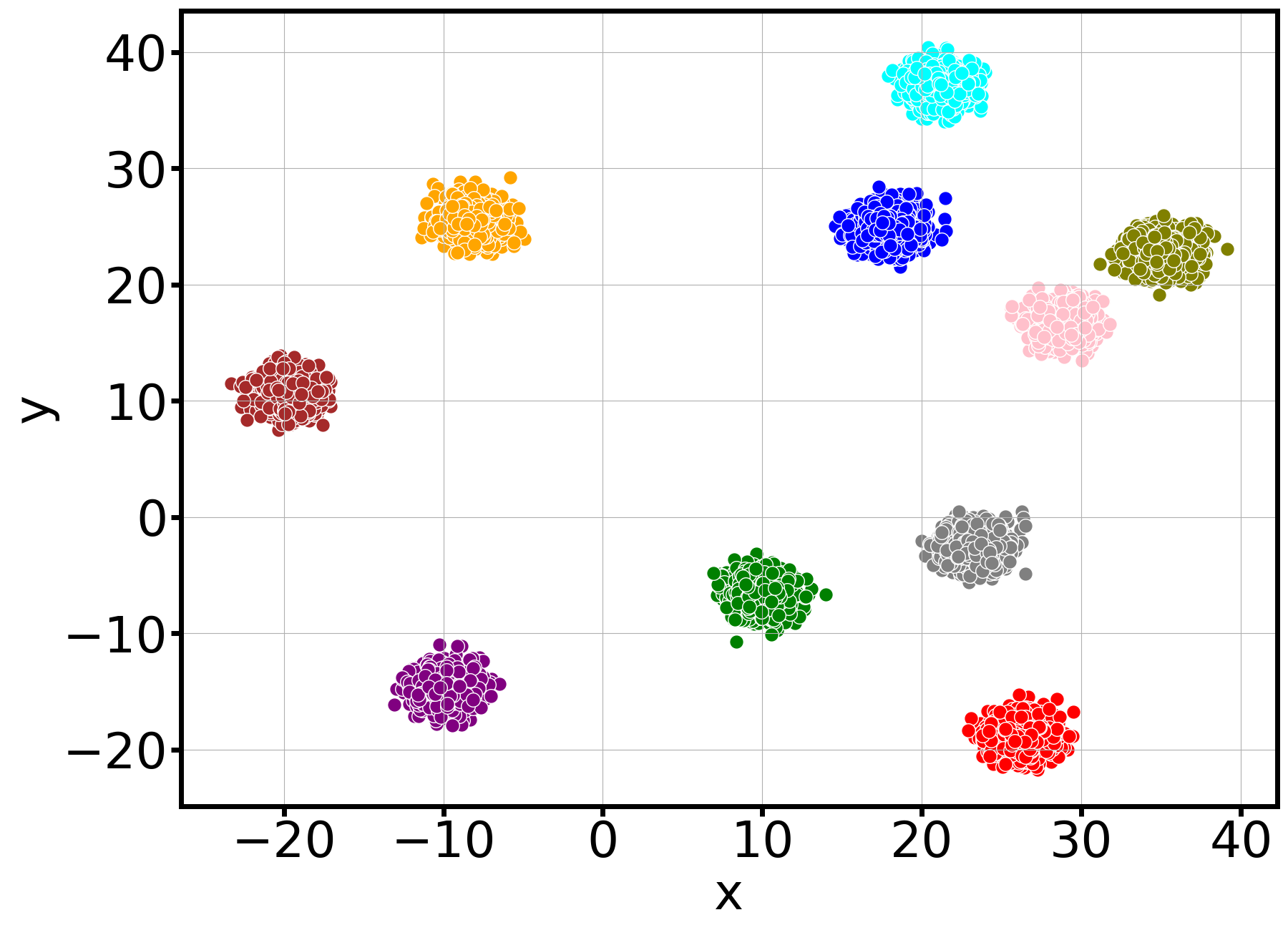}}\hspace{16pt}
  \subfloat[Digits dataset\label{fig:digits}]{\includegraphics[width=0.46\linewidth]{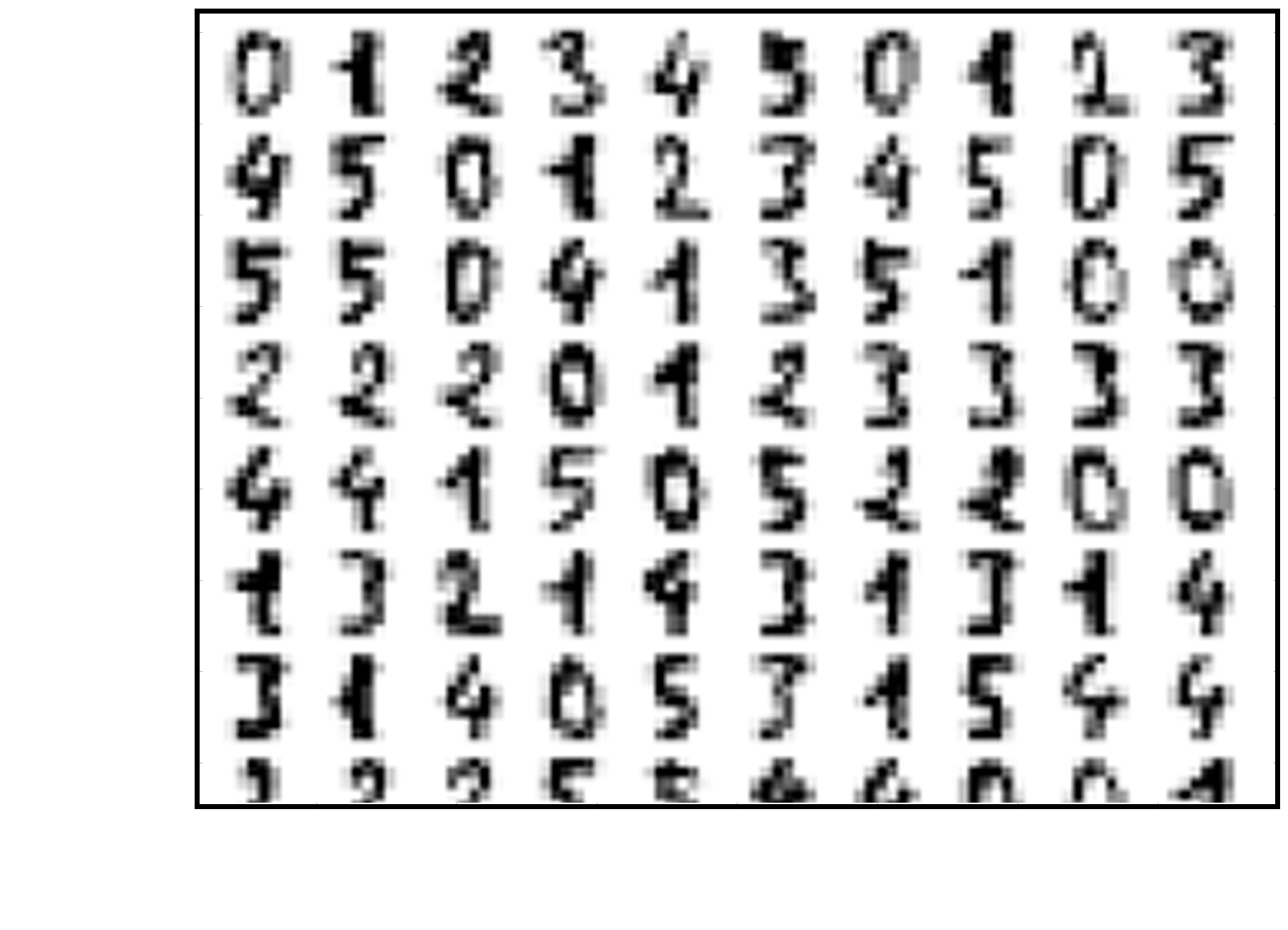}}\\[8pt]
  \subfloat[Network Architectures\label{tab:architectures}]{
    \begin{tabular}{cccc} \toprule 
      Architecture & \texttt{moons}, \texttt{circles} & \texttt{blobs} & \texttt{digits} \\ \midrule
      \small A & \footnotesize [2, 20, 2] & \footnotesize [2, 20, n] & \footnotesize [64, 20, n] \\
      \small B & \footnotesize [2, 75, 2] & \footnotesize [2, 75, n] & \footnotesize [64, 75, n] \\
      \small C & \footnotesize [2, 100, 2] & \footnotesize [2, 100, n] & \footnotesize [64, 100, n] \\
      \small D & \footnotesize [2, 50, 50, 2] & \footnotesize [2, 50, 50, n] & \footnotesize [64, 50, 50, n] \\ \bottomrule
    \end{tabular}}
    \caption{Overview of the datasets and network architectures used in our study: (\subref{fig:moons}) The \texttt{moons} test dataset, comprising 16,000 two-dimensional data points normalized within the range $[-0.7, 0.7]$, and (\subref{fig:circles}) the \texttt{circles} test dataset, featuring two concentric rings with 16,000 two-dimensional data points. (\subref{fig:blobs}) We generated 10 randomly distributed clusters consisting of 12,500 data points for the two-dimensional \texttt{blobs} test dataset and (\subref{fig:digits}) shows a random subset of samples from the \texttt{digits} dataset. All datasets have been generated using methods provided by \texttt{scikit-learn} \citep{pedregosa2011scikit}. The network architectures (\subref{tab:architectures}) are identified by single-letter codes. The bracket notation represents the number of neurons in each layer, where the first value indicates the number of input neurons, and the last value denotes the number of output neurons. For the \texttt{blobs} and \texttt{digits} dataset, the number of output neurons \texttt{n} depends on the number of classes used. Figures and Table adapted from \cite{ncta24}.
    }
\end{figure*}

\paragraph{Datasets}
The experiments were conducted on the basis of four different data sets, with different degrees and sources of complexity.
Classification problems were selected for the experiments as they are characterized by their interpretability and the availability of a well-established evaluation metric that facilitates objective analysis. As illustrated in Fig.~\ref{fig:moons}, the two-dimensional \texttt{moons} dataset comprises two moon-shaped point clusters with minimal overlap, classified into only two categories. A neural network with one hidden layer consisting of 6 hidden units, and trained via backpropagation, attains approximately $100\%$ accuracy in select runs.
\\[2pt]
Contrasting with this relatively simple dataset, we have chosen the more complex 2D binary classification problem represented by the \texttt{circles} dataset, as shown in Fig.~\ref{fig:circles}. This data set comprises two different classes that form two concentric circles. The larger ring is arranged in such a way that it encloses the smaller ring. The transition from the smaller to the larger ring is instantaneous, and there are numerous points in the overlap region. This property poses a significant challenge even for the trained dense network and emphasizes the complexity of the problem at hand. A total of $66,000$ random data points were generated for both datasets. Subsequently, Gaussian noise with a standard deviation of $0.07$ was added to each dataset. To test the GA's performance in multi-class classification scenarios, we artificially generated the \texttt{blobs} dataset, which comprises up to 10 distinct two-dimensional Gaussian-shaped clusters, being assigned class labels $\{0, ..., 9\}$. These clusters are uniformly dispersed in the feature space, ensuring they do not overlap, as depicted in Fig.~\ref{fig:blobs}. Networks optimized with the backpropagation algorithm have been observed to achieve 100\% accuracy in the classification of points within the two-dimensional space, irrespective of the number of distinct classes (cf. Fig.~\ref{fig:blobs_alg_comp}). As a final benchmark, we included the \texttt{digits} dataset (cf. Fig.~\ref{fig:digits}), which is a lower-dimensional version similar to the popular MNIST dataset for classifying handwritten digits \cite{6296535}. It consists of $1797$ images with $8 \times 8$ input features corresponding to individual pixels, and are assigned class labels $\{0, ..., 9\}$.  
The datasets were split into a training set and a test set, with 25\% of the data points designated for evaluation. Furthermore, to circumvent the potential adverse scaling effects that might originate from non-Gaussian distributions, a min-max normalization procedure was applied to the \texttt{moons} and the \texttt{digits} datasets.

\paragraph{Network Architectures}
We employ feed-forward ANNs with rectified linear unit activation functions (ReLU) exclusively for the neurons in the input and hidden layers. When we use accuracy as performance objective, we do not incorporate a softmax activation function, since we can directly compute accuracies without requiring class probabilities, by determining the class of the network output with the highest value. To gain insight into the behavior of the GA across various model sizes, four network architectures are examined in the experimental section (Table~\ref{tab:architectures}). To streamline the presentation of the subsequent plots, the analyzed network architectures are denoted by ``A'', ``B'', ``C'', and ``D'' for convenience. 
Our studies demonstrated that the selection of the network parameter initialization method has a substantial influence on the final accuracy achieved. In our experiments using the genetic algorithm, we uniformly sampled the network weights from the interval $[-1, 1]$. This approach consistently yielded the most optimal results across all examined datasets. These findings are supported by theoretical proofs for the existence of SLTs in uniformly initialized networks \citep{malach2020proving, pensia2020optimal}. Notably, while the majority of research in the area of SLTH employs zero biases, our experiments revealed a substantial improvement in performance when biases were initialized based on the same uniform distribution. 

\paragraph{Baselines}
Since strong lottery tickets are defined based on their comparative performance to networks obtained by classical gradient-based training, we use backpropagation as a baseline. To benchmark against a well-optimized implementation of a feed-forward network and its trainer, we employed \textit{scikit-learn’s} \textit{MLPClassifier} and conducted hyperparameter tuning across all $4$ architectures using their \textit{RandomizedSearchCV} method. We choose random search for its computational efficiency in navigating large parameter spaces with limited resources. The parameter ranges were defined based on prior insights and preliminary trials. Tuned hyperparameters include solvers, learning rates, batch sizes, momentum, L2 regularization strengths (alphas), and epsilon values for numerical stability. Table \ref{tab:backprop_hyperparameters} summarizes the selected values. To ensure convergence, both the search and training ran for $1000$ epochs. Our study evaluates the mean test accuracies of backpropagation-trained networks listed in Table \ref{tab:architectures}.

\begin{table}[h]
\centering
\resizebox{\columnwidth}{!}{%
\begin{tabular}{ccccccccc}
\toprule
Dataset & Classes & Solver & Learning Rate & Learning Rate Init & Epsilon & Batch Size & Alpha & Momentum \\
\midrule
\multirow{4}{*}{moons (A-D)} 
& 2 & adam & constant & 0.021544 & 4.64e-09 & 128 & 0.0001 & - \\
& 2 & adam & constant & 0.001 & 4.64e-09 & 64 & 0.000215 & - \\
& 2 & adam & constant & 0.001 & 4.64e-09 & 64 & 0.000215 & - \\
& 2 & adam & constant & 0.001 & 4.64e-09 & 64 & 0.000215 & - \\
\midrule
\multirow{4}{*}{circles (A-D)} 
& 2 & sgd & adaptive & 0.1 & - & 64 & 0.046416 & 0.0 \\
& 2 & sgd & adaptive & 0.004642 & - & 128 & 0.046416 & 0.5, nesterov \\
& 2 & adam & constant & 0.001 & 4.64e-09 & 64 & 0.000215 & - \\
& 2 & sgd & adaptive & 0.1 & - & 128 & 0.046416 & 0.0, nesterov \\
\midrule
\multirow{1}{*}{blobs (C)}
& 2-10 & adam & constant & 0.001 & 4.64e-09 & 64 & 0.000215 & - \\
\midrule
\multirow{5}{*}{digits (B)}
& 2 & adam & constant & 0.001 & 4.64e-09 & 64 & 0.000215 & - \\
& 3 & adam & constant & 0.021544 & 1.67e-09 & 128 & 0.046416 & - \\
& 4 & adam & adaptive & 0.002783 & 7.74e-09 & 32 & 0.004642 & - \\
& 5 & adam & adaptive & 0.002783 & 5.99e-08 & 32 & 0.01 & - \\
& 10 & adam & adaptive & 0.002783 & 7.74e-09 & 32 & 0.004642 & - \\
\bottomrule
\\
\end{tabular}
}
\caption{Listing of the determined backpropagation hyperparameters for the MLPClassifier model from \textit{scikit-learn} using random search. The letters in brackets correspond to our 4 network architectures. For the two multi-class datasets, we only consider two architectures in our experiments. Table adapted from \cite{ncta24}.}
\label{tab:backprop_hyperparameters}
\end{table}

\section{Experimental Results}
This section provides a comprehensive evaluation of the GA's capabilities in finding strong lottery tickets. Section~\ref{subsec:GA_perf} investigates the average performance of our two GA configurations on the \texttt{moons} and \texttt{circles} datasets, followed by a general analysis regarding runtime complexity. Section~\ref{subsec:Edge_popup_perf} contains an ablation study for the state-of-the-art edge-popup, where we analyze the impact of different weight initializations on the algorithms' performance. Based on our results, we conduct a broad performance study in which we compare the accuracy of the subnets found by our GA with that found by edge-popup and the backpropagation-trained networks. We employ a mixed linear model, to quantify the statistical significance of our results. Finally, in Section~\ref{subsec:prob_multiclass}, we test our best GA configuration on our multi-class classification tasks, using accuracy and cross-entropy loss as performance metrics for evaluating our individuals. We analyze how the choice of the performance objective affects the complexity of the optimization, as well as how important the normalization of datasets is. We conclude by highlighting the additional sparsity gains we achieve by using the post-evolutionary pruning routine.

\subsection{GA Performance Analysis}\label{subsec:GA_perf}
\begin{figure*}[t]\centering
\subfloat[\texttt{moons}, $R = 50$\label{fig:GA_moons}]{\includegraphics[width=0.28\linewidth]{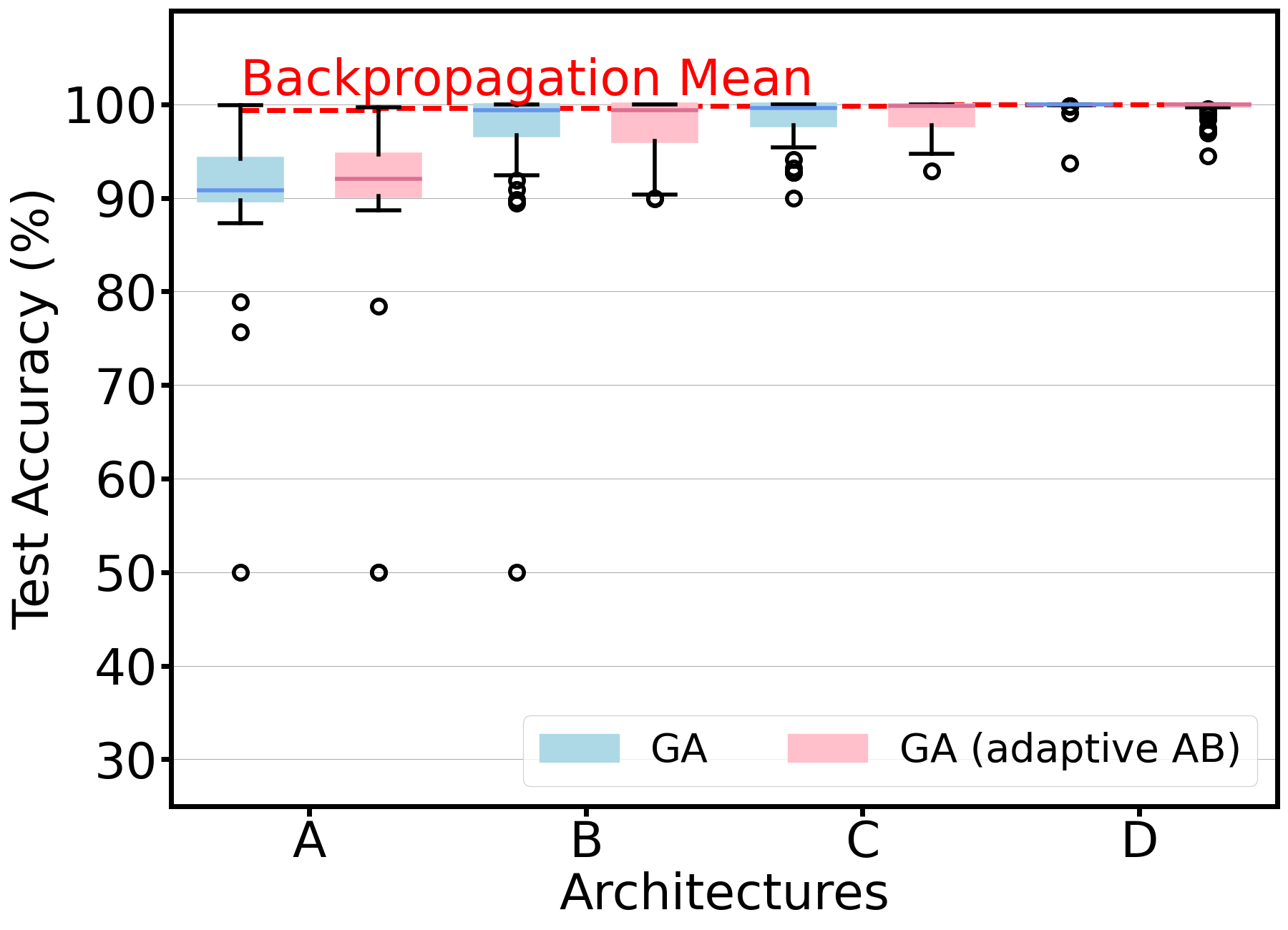}}
\subfloat[\texttt{circles}, $R = 50$ \label{fig:GA_circles}]{\includegraphics[width=0.28\textwidth]{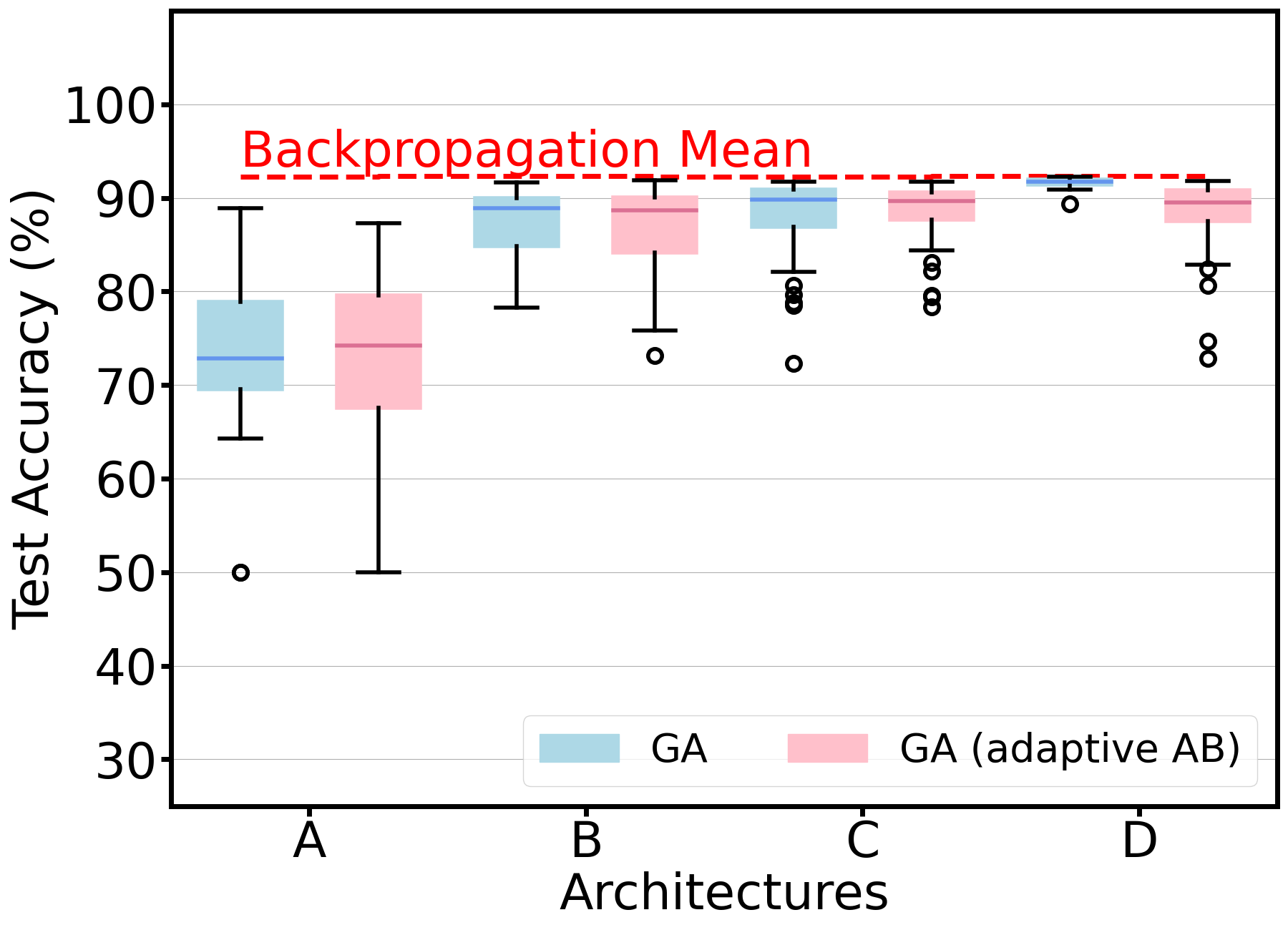}}
\subfloat[Reached mean accuracies $\pm$ standard deviation\label{tab:GA_accuracies}]{
\raisebox{48pt}[0pt][0pt]{
\resizebox{0.4\linewidth}{!}{
\tiny
\begin{tabular}{ccccc}\toprule
 & Arch. & GA & GA (adaptive AB) & Backprop.\\
 \midrule
 \parbox[t]{2mm}{\multirow{4}{*}{\rotatebox[origin=c]{90}{\texttt{moons}}}} 
 & A & 90.7\% $\pm$ 7.2 & 90.9\% $\pm$ 9.1 & 99.4\% $\pm$ 2.4 \\
 & B & 96.9\% $\pm$ 7.3 & 97.5\% $\pm$ 3.3 & 99.6\% $\pm$ 1.9 \\
 & C & 98.4\% $\pm$ 2.5 & 98.9\% $\pm$ 1.7 & 99.8\% $\pm$ 1.4 \\
 & D & 99.8\% $\pm$ 0.9 & 99.6\% $\pm$ 1.0 & 99.9\% $\pm$ 0.0 \\
 \midrule
\parbox[t]{2mm}{\multirow{4}{*}{\rotatebox[origin=c]{90}{\texttt{circles}}}} 
 & A & 73.9\% $\pm$ 8.1 & 73.7\% $\pm$ 8.6 & 92.3\% $\pm$ 0.1 \\
 & B & 87.3\% $\pm$ 3.6 & 86.8\% $\pm$ 4.4 & 92.4\% $\pm$ 0.0  \\
 & C & 88.0\% $\pm$ 4.3 & 88.5\% $\pm$ 3.2 & 92.3\% $\pm$ 0.0  \\
 & D & 91.6\% $\pm$ 0.5 & 88.3\% $\pm$ 4.0 & 92.4\% $\pm$ 0.0  \\
\bottomrule
\vspace{-8mm}
\end{tabular}}}}
\caption{Summary of the GA performance on the \texttt{moons}~(\subref{fig:GA_moons}) and \texttt{circles}~(\subref{fig:GA_circles}) datasets. The blue boxplots represent multiple runs for each architecture using the default GA configuration, while the pink boxplots show the results of $R$ runs with the GA configuration that incorporates an adaptive accuracy bound, initialized with a threshold value of $0.85$. For reference, we include the mean accuracies obtained from trained networks using backpropagation. The achieved accuracies are detailed in (\subref{tab:GA_accuracies}). Figures and Table taken from \cite{ncta24}.}
\label{fig:GA_perform}
\end{figure*}

As previously stated, our experiments utilize four distinct network architectures (see Table~\ref{tab:architectures}). We hypothesize that networks with more parameters are more likely to include favorable initializations, potentially yielding higher-performing subnetworks. Furthermore, we investigate whether applying an accuracy abound when generating the initial population influences the subsequent evolutionary process. In this section, we only consider binary classification problems, and we use accuracy as the performance objective when evaluating the fitness of individuals.
\\[2pt]
As illustrated in Fig.~\ref{fig:GA_moons}, the outcomes of the \texttt{moons} dataset indicate that the genetic algorithm consistently achieves high final accuracies for network D, attaining a mean accuracy approaching 100\%. Upon examining the variability among GA runs across different network architectures, a noticeable correlation emerges between the quantity of network parameters and the resulting performance. For the least complex network A, which contains only 80 parameters, the mean difference to backpropagation is approximately 9\%. As the parameter count increases, the discrepancy between the two methods decreases steadily. For networks C and D, the mean value approaches that of backpropagation, and for network D, there is minimal variance between the runs.There is less variability in performance between the distinct GA configurations. Overall, the mean for the runs using an accuracy bound seem to exhibit slight increases relative to the mean from runs not employing an accuracy bound, though this effect diminishes with increasing network sizes. 

\begin{figure*}[t]\centering
    \subfloat[Accuracy\label{fig:acc_dev}]{\includegraphics[width=0.46\textwidth]{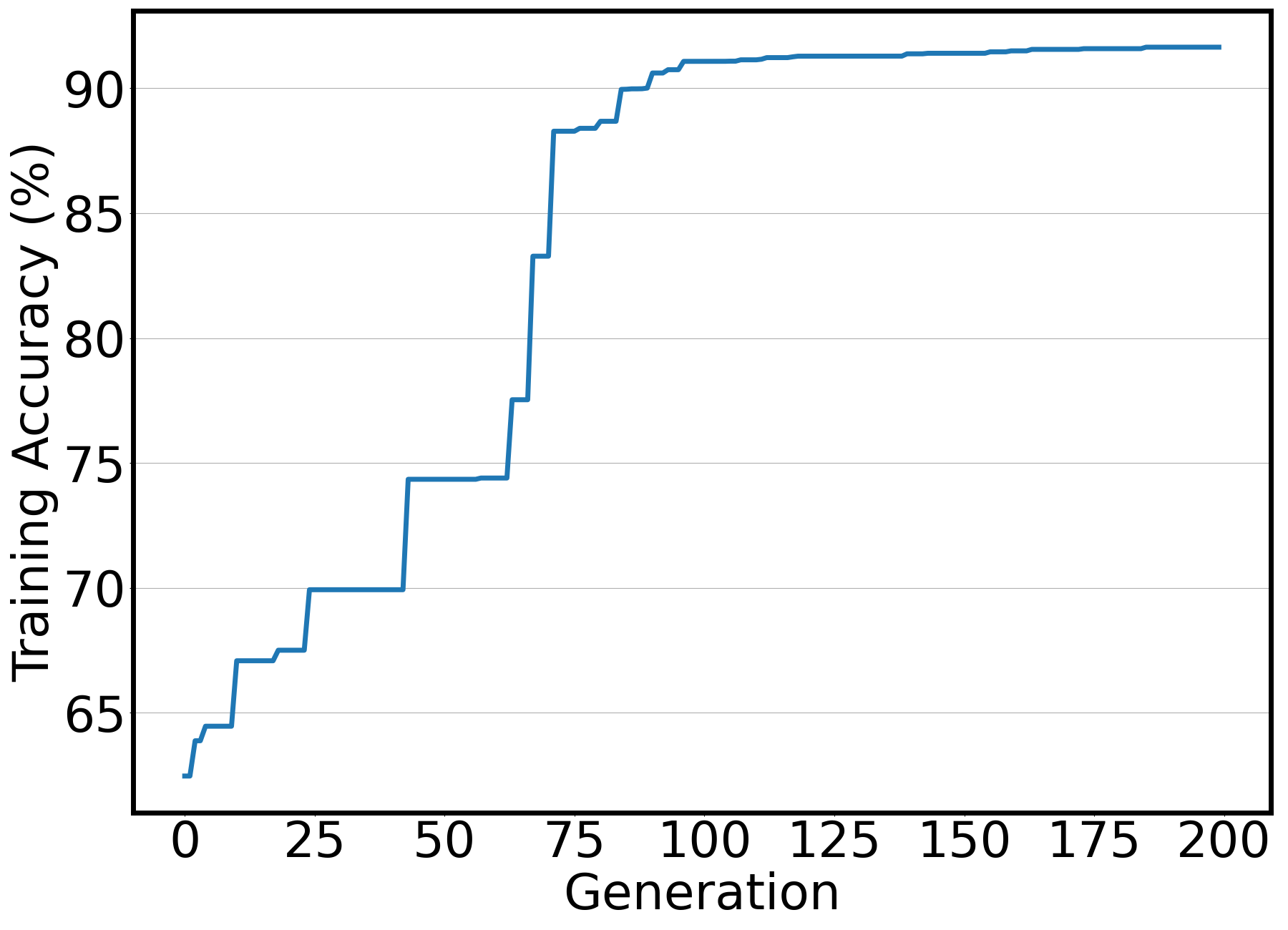}}\hspace{0.05\textwidth}
    \subfloat[Sparsity\label{fig:spars_dev}]{\includegraphics[width=0.46\textwidth]{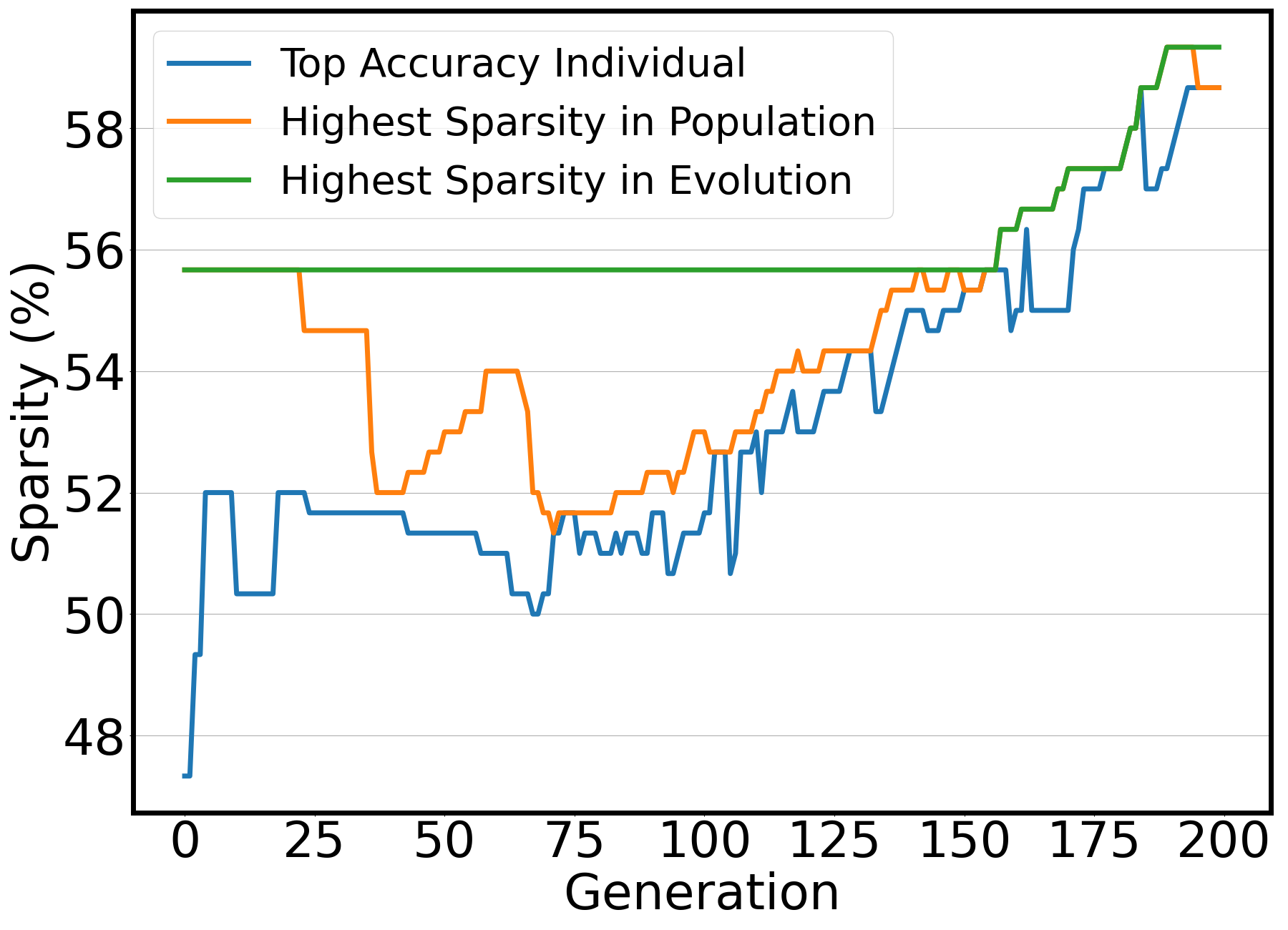}}
    \caption{Optimization progress of a successful run using ``GA (adaptive AB)'' on network architecture B $=[2, 75, 2]$ for the \texttt{circles} dataset, tracking accuracy (\subref{fig:acc_dev}) and sparsity (\subref{fig:spars_dev}). The blue line represents the sparsity of the fittest individual in the current population, the orange line indicates the highest sparsity within the current generation, and the green line marks the best sparsity found across all previous generations. Figures taken from \cite{ncta24}.}
    \label{fig:GA_example_run}
\end{figure*}

\noindent The results shown in Fig.~\ref{fig:GA_circles} are largely consistent with the previous findings. A comparison of the mean performance of backpropagation across the two datasets reveals higher complexity levels for the \texttt{circles} dataset. The GA consistently achieves the lowest accuracies on network architecture A, but achieves higher final accuracies on larger networks. The usage of network D resulted in the highest mean accuracy, with a result of $91.6 \%$. Notably, this outcome was attained without the employment of an accuracy bound, and there also seems to exist a minimum parameter count threshold below which final accuracies are significantly lower. Conversely, a similar positive effect as the network size increases is less pronounced. Overall, the results suggest that there are situations where the GA is able to achieve accuracies similar to backpropagation.  
\\[2pt]
To analyze the typical behavior of the GA in optimizing accuracy and sparsity, we examined a high-performing run from the experiments on the \texttt{circles} dataset. This run was conducted using network architecture B and the GA configuration with an adaptive accuracy bound. As shown in Fig.~\ref{fig:acc_dev}, the fittest individual in the initial population achieved less than $65\%$ accuracy. Over the course of the first 100 generations, accuracy improved significantly, with rapid jumps, before stabilizing at around $91\%$, demonstrating the GA’s optimization capabilities. After this point, until the termination condition is satisfied (i.e. reaching 200 generations), only marginal improvements occur.
\\[2pt]
Fig.~\ref{fig:spars_dev} illustrates the evolution of sparsity throughout the same run. Initially, sparsity decreases as accuracy is prioritized. Once accuracy gains slow down, the GA begins optimizing sparsity more effectively. At this stage, the population becomes highly homogeneous, with many individuals achieving similar accuracy levels. Consequently, the GA shifts focus to sparsity, ultimately improving it by approximately $10\%$ compared to the best individual in the initial population. Note that this graph only shows the sparsity development throughout the evolution and does not include the final sparsity level obtained after employing our post-evolutionary pruning routine. 

\paragraph{Scalability}\label{subsec:scalability}
The calculation of the fitness value for each individual in the population is the critical factor in determining the runtime complexity of our GA. It can be expressed as the function $\mathcal{O}(g*N*(d*l*b^{2}))$ describing the total number of multiplications performed during evolution, with the number of generations $g$, the population size $N$, the number of dataset samples $d$ and the worst case network architecture characterized by $l*b^{2}$ parameters (that is, the length of the bit-vector). Typically, it holds that $N < g$ and $(l*b^{2}) \ll d$. In practice, the impact of population size and number of generations on runtime can be mitigated by implementing efficient parallelization. Furthermore, a compressed version of the subnetwork encoding can reduce the complexity of the other GA operations.

\subsection{Edge-Popup \& Weight Initialization}\label{subsec:Edge_popup_perf}
In the preceding subsection, we demonstrated that the GA performs effectively on binary classification tasks, achieving accuracies comparable to or even matching those obtained through backpropagation, provided an adequate network architecture is selected. To assess how the GA compares to other methods for discovering SLTs in randomly initialized neural networks, we repeat our experiments using the well-known edge-popup algorithm \citep{ramanujan2020s}.
\\[2pt]
Edge-popup assigns a score to each network weight and extracts subnetworks by selecting the top $k\%$ scoring edges per layer for the forward pass. These scores are adjusted during backpropagation using the straight-through gradient estimator \citep{bengio2013estimating}. Unlike static pruning, edge-popup allows previously removed edges to reappear, as their contributions to the loss are continuously reevaluated during gradient approximation. The parameter $k$, referred to as the pruning rate, defines the fraction of weights retained. For instance, a $60\%$ pruning rate results in a subnetwork where $(1 - k) = 40\%$ of weights are pruned. It is important to note that our sparsity metric is defined oppositely: a sparsity of $60\%$ corresponds to 60\% of the weights being pruned.
\\[2pt]
We follow the authors' default settings and train for a total of $100$ epochs, evaluating each configuration across $25$ random seeds. Their experiments identified two particularly effective initialization methods: initializing network parameters from a Kaiming normal distribution (also referred to as He initialization \citep{he2015delving}), which we denote as ``Weights $\sim \mathcal{N}_{k}$'' following Ramanujan et al. \citep{ramanujan2020s}, and sampling from a signed Kaiming constant distribution, referred to as ``Weights $\sim U_{k}$''.

\begin{figure*}[t]\centering
    \subfloat[\texttt{moons, $R = 25$}\label{fig:edge_moons}]{\includegraphics[width=0.46\textwidth]{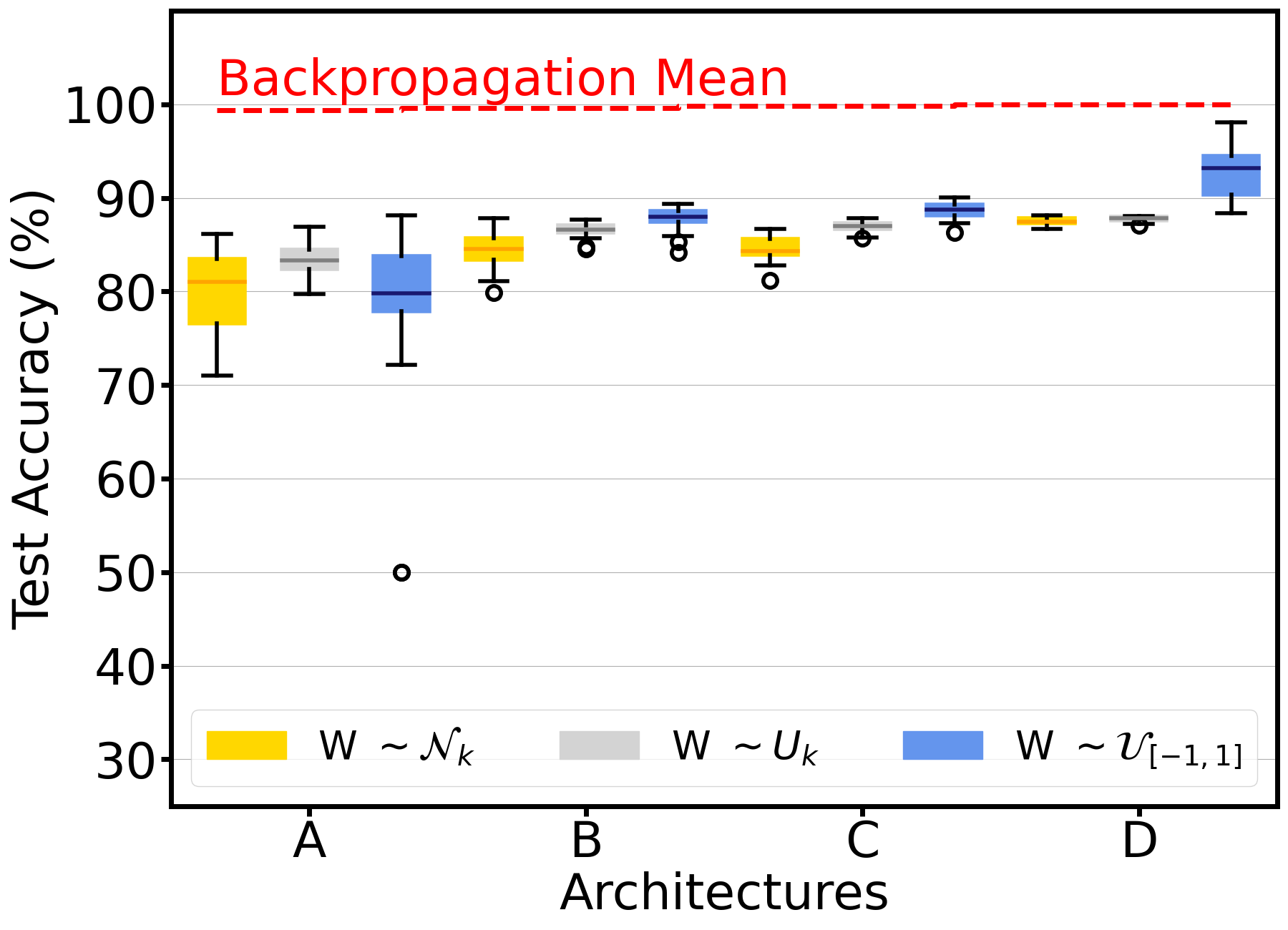}}\hspace{0.05\textwidth}
    \subfloat[\texttt{circles, $R = 25$}\label{fig:edge_circles}]{\includegraphics[width=0.46\textwidth]{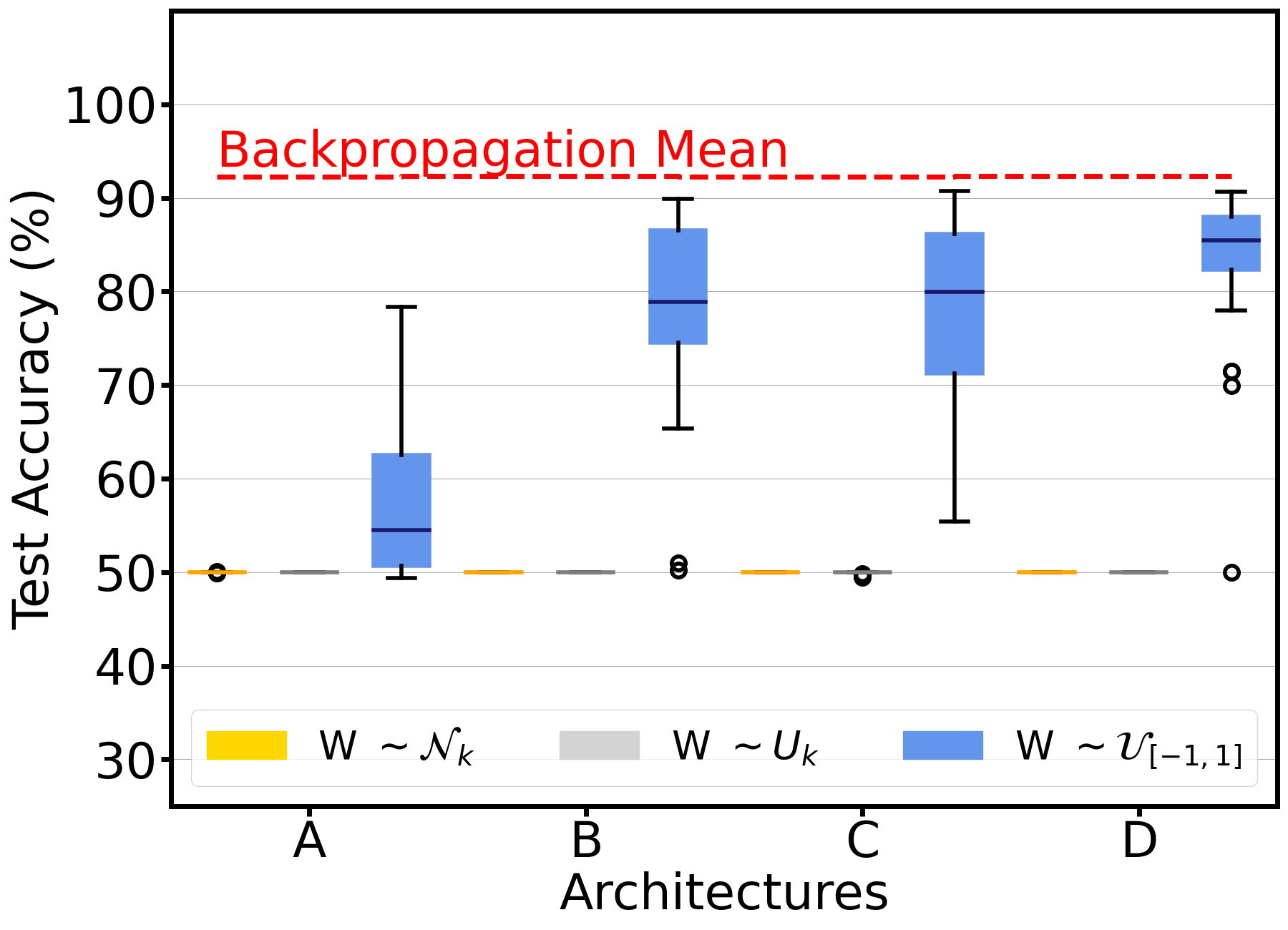}}
    \caption{Visualization of edge-popup's performance across the displayed datasets, using different color-coded initializations with $R$ runs each. For comparison, the mean accuracies achieved through backpropagation on the respective architectures are shown as dashed lines. Figures taken from \cite{ncta24}.}
    \label{fig:edge_perform}
\end{figure*}

\noindent In addition to our own initialization method, labeled as ``Weights $\sim \mathcal{U}_{[-1, 1]}$'' we also include runs where networks are initialized using both of their approaches. Notably, we employ the scaled versions of these methods, where the standard deviation is adjusted by a factor of $\sqrt{1/k}$. For precise definitions, refer to Ramanujan et al. \citep{ramanujan2020s}. Analogously to our GA setup, when using our parameter initialization method with edge-popup, we sample biases from a uniform distribution.
\\[2pt]
Given the significant impact of parameter initialization on our GA’s performance —consistent with the findings of Ramanujan et al. \citep{ramanujan2020s}—we begin with an ablation study to identify the initialization technique that yields the highest accuracy for edge-popup before conducting the main comparison experiments. Fig.~\ref{fig:edge_moons} presents the results of edge-popup runs on the \texttt{moons} dataset. A clear trend emerges for the impact of the network size on the final accuracies, akin to the one we observed for the GA: The larger networks generally achieve higher final accuracies on the \texttt{moons} dataset. Additionally, apart from network A, runs using our initialization method consistently outperformed the others. Notably, for network D, our method achieved a substantial improvement of $\approx 5\%$ in mean accuracy compared to the other configurations. However, in all cases, edge-popup’s mean accuracy remains noticeably lower than that of backpropagation.
A similar pattern emerges in the \texttt{circles} experiment, as illustrated in Fig.~\ref{fig:edge_circles}. However, in this case, the Kaiming normal and signed Kaiming constant distributions proved entirely ineffective. None of the runs achieved classification accuracy beyond random guessing, only predicting the correct label in 50\% of cases. At first glance, one might suspect an issue with the edge-popup algorithm itself. However, since it performs well with our initialization method, the problem most likely lies within the usage of the Kaiming normal and signed Kaiming constant distributions. A possible explanation is that the Gaussian nature of the rings in the dataset introduces distortions that negatively impact these initialization methods. Determining the exact cause remains an open question for future research.
In summary, the results indicate that edge-popup and the GA both benefit from the usage of a uniform initialization for the \texttt{moons} and \texttt{circles} datasets. We therefore decided to concentrate on the ``Weights $\sim \mathcal{U}_{[-1, 1]}$'' configuration for the subsequent extensive comparative analysis. 
\\[2pt]
Given the inverse relationship between accuracy and sparsity in the early stages of evolution (cf. Fig.~\ref{fig:spars_dev}), we hypothesize a correlation between these factors. This suggests that the number of parameters pruned throughout evolution may influence the final fitness achieved. Unlike our approach, edge-popup keeps the number of pruned connections in the subnetworks constant. To ensure that its performance is not hindered by this strictness, we include additional edge-popup runs in our comparison study, where we set the pruning rates to match the mean sparsity levels achieved by the two GA configurations. For each architecture and dataset, we selected the GA configuration that achieved the highest mean accuracy and reran the edge-popup experiments using the corresponding mean sparsity levels.

\begin{figure*}[p]\centering
  \subfloat[\texttt{moons}\label{fig:edge_moons_adapted}]{\includegraphics[width=0.46\textwidth]{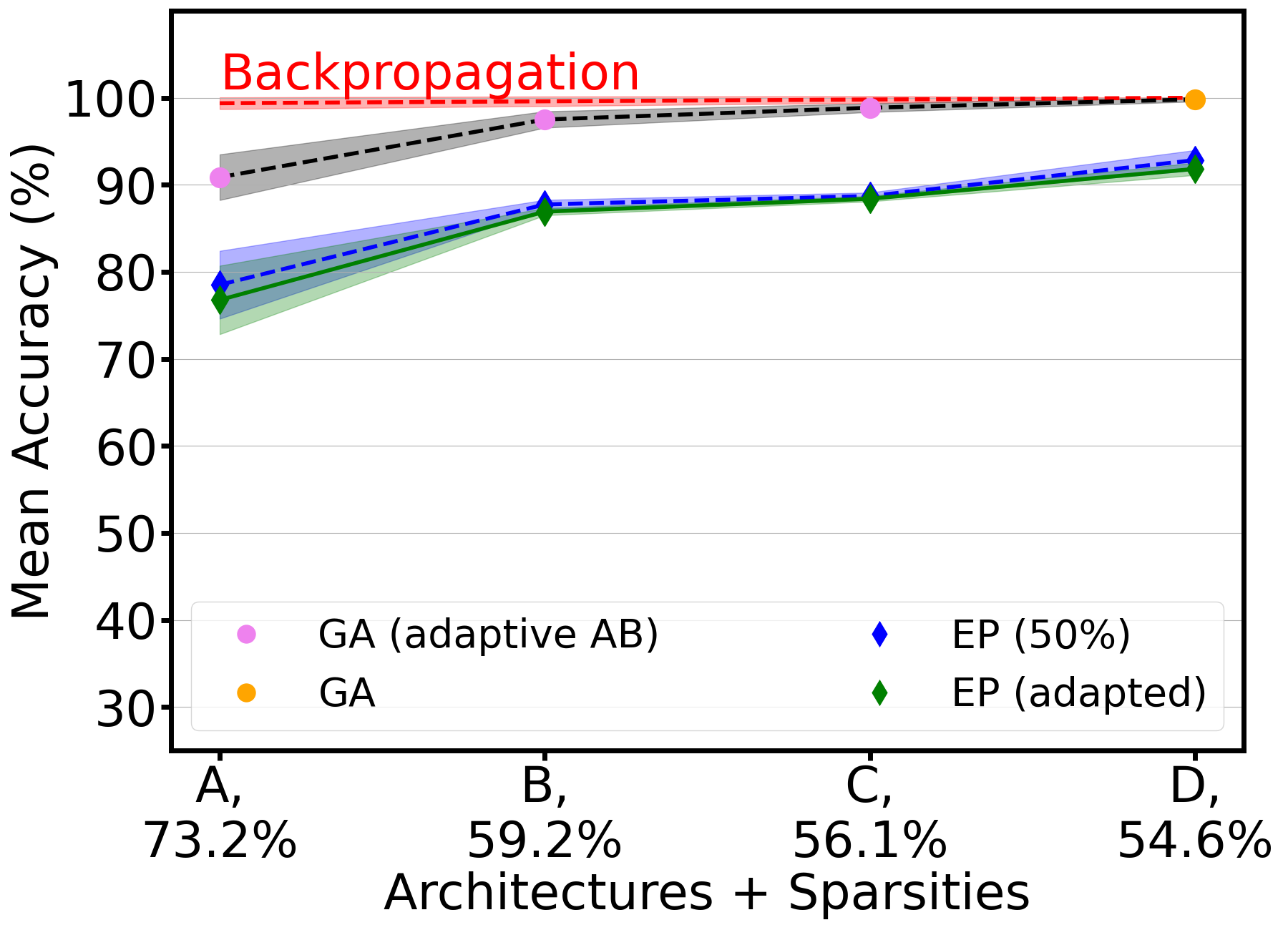}}
  \hspace{0.05\textwidth}
  \subfloat[\texttt{circles}\label{fig:edge_circles_adapted}]{\includegraphics[width=0.46\textwidth]{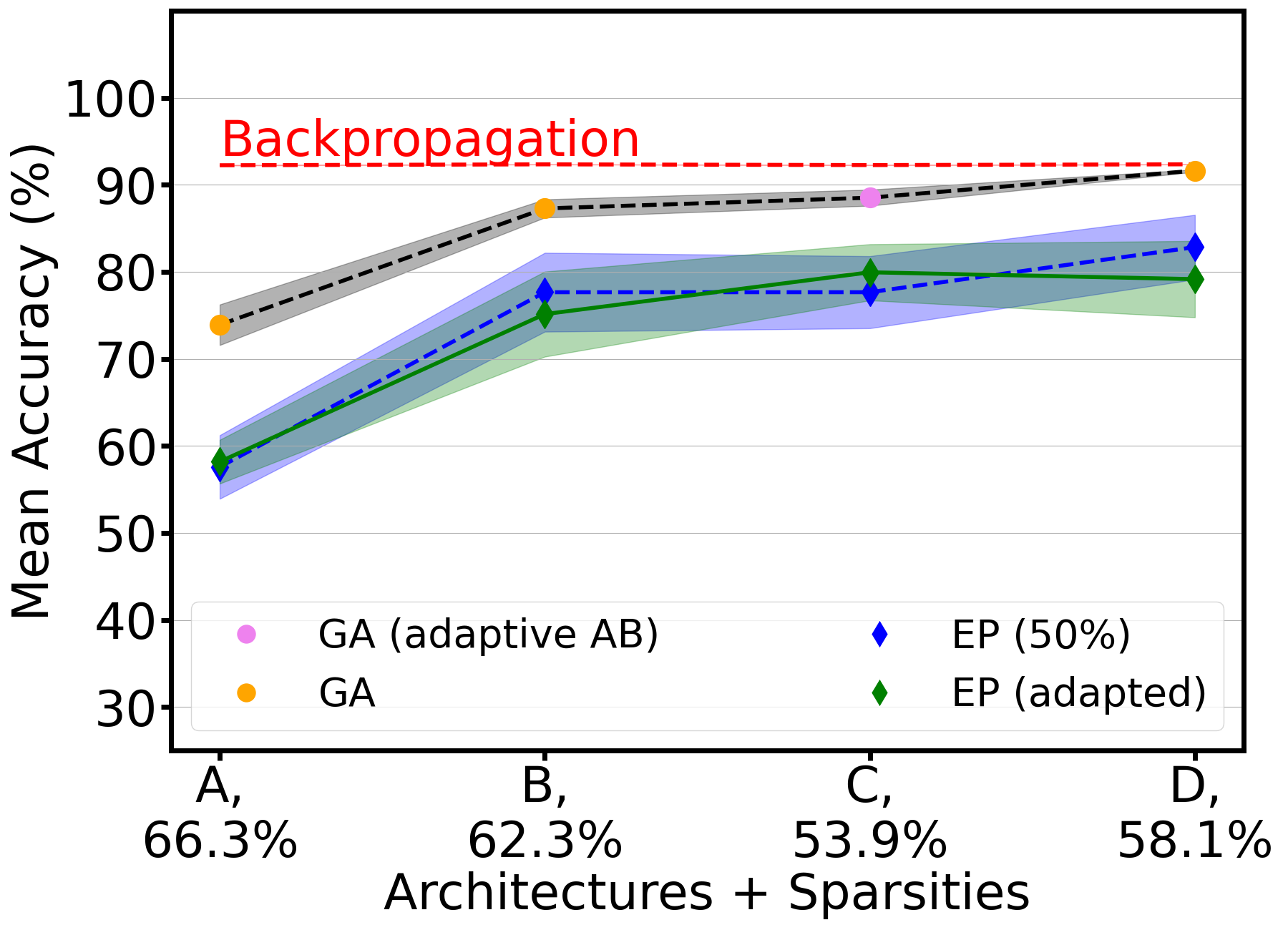}}\\[8pt]
  \subfloat[Statistical analysis\label{tab:statistics}]{\centering
    \small\begin{tabular}{cccccccc}
        \toprule
        Dataset & Reference & Target & Coef. & Std. Err. & $z$ & $P > |z|$ & 95\%-Conf. \\
        \midrule
        \multirow{4}{*}{moons} & GA & GA (adaptive AB) & 0.042 & 0.083 & 0.507 & 0.612 & [-0.121, 0.205] \\
         & EP (50\%) & EP (adapted) & -0.133 & 0.095 & -1.405 & 0.160 & [-0.320, 0.053] \\
         & GA & EP (50\%) & -1.185 & 0.081 & -14.697 & 0.000 & [-1.343, -1.027] \\
         & GA & Backpropagation & 0.661 & 0.087 & 7.609 & 0.000 & [0.491, 0.831] \\
        \midrule
        \multirow{4}{*}{circles} & GA & GA (adaptive AB) & -0.106 & 0.064 & -1.670 & 0.095 & [-0.231, 0.018] \\
         & EP (50\%) & EP (adapted) & -0.062 & 0.102 & -0.609 & 0.543 & [-0.263, 0.138] \\
         & GA & EP (50\%) & -0.964 & 0.074 & -13.095 & 0.000 & [-1.109, -0.820] \\
         & GA & Backpropagation & 1.030 & 0.071 & 14.595 & 0.000 & [0.892, 1.168] \\
        \bottomrule
    \end{tabular}} 
    \caption{Comparison of GA and edge popup performance across the displayed datasets using the sparsity levels achieved by our GA configurations as new fixed pruning rates for edge-popup. The selected mean sparsity levels are from either ``GA'' or ``GA (adaptive AB)'', as indicated by the different colored dots, depending on their respective mean accuracies. For reference, we show the mean accuracies and 95\% confidence intervals for the corresponding GA configuration, backpropagation, and the original edge popup variant with the default pruning rate of $0.5$. A final statistical analysis evaluates the performance differences between the algorithm combinations and provides p-values for the GA and edge popup configurations as well as for the backpropagation baseline. Figures and Table taken from \cite{ncta24}.}
    \label{fig:edge_perform_adapted}
\end{figure*}

\noindent Fig.~\ref{fig:edge_perform_adapted} presents the results of our comparison study. To provide a comprehensive evaluation, we plotted the mean accuracy of the best-performing GA configuration for each architecture, alongside the mean accuracies of backpropagation and the original edge-popup runs. The shaded regions around the line plots indicate the 95\% confidence intervals for the mean estimates. To facilitate the comparison of edge-popup with the adjusted pruning rates, we also display the mean sparsity levels achieved by our GA configurations (during the evolution) on the x-axis, below the network architectures. As shown in Fig.~\ref{fig:edge_moons_adapted}, these sparsity levels decreased as network size increased, eventually converging to 0.5, which coincides with edge-popup’s default pruning rate. This suggests that the impact of varying the pruning rate should be more pronounced in smaller architectures.
\\[2pt]
Indeed, the most significant relative change is observed in network A. Here the adjusted pruning rate led to several low-accuracy runs, lowering the overall mean. However, due to high variance, some instances still outperformed \textit{EP (50\%)}. For the larger networks, the mean accuracy remained largely unchanged, and any observed differences were negative.
A similar pattern is observed for the \texttt{circles} dataset, as shown in Fig.~\ref{fig:edge_circles_adapted}, with the exception of architecture C. Given the substantial variance among runs with the same pruning rate and the largely overlapping confidence intervals, it is unclear whether these changes can be attributed to the adjusted pruning rates. Overall, none of the modifications resulted in a notable performance improvement.
\\[2pt]
When comparing edge-popup to the GA configurations, it is clear that the GA consistently outperforms edge-popup across all datasets and architectures, even when edge-popup is adapted to discover sparser subnetworks than in its original version. In fact, the adjusted pruning rates result in worse performance. From this, we can conclude that the GA is capable of finding subnetworks with higher accuracy that are also sparser, and that, for larger networks, it achieves performance comparable to backpropagation. 
To evaluate the statistical significance of our findings, we apply a linear mixed model to the accuracy data, particularly focusing on the comparison of various algorithms across different architectures within a shared dataset. In doing so, we treated the algorithms as fixed effects, while we modeled the four architectures and different network initializations as random effects, thereby aiming to capture variability across runs. We perform our statistical analysis using the \textit{MixedLM} module from \citep{seabold2010statsmodels}. In order to properly fit the model and ensure convergence, we use  \textit{Powell's algorithm}, along with restricted maximum likelihood (REML), and standardize the accuracy values. The results of this analytical process are outlined in Table \ref{tab:statistics}.
\\[2pt]
To evaluate statistical significance, we analyze various statistics, including coefficients and p-values, to understand the relationship between the reference and target algorithms. For the \texttt{moons} dataset, the positive coefficient for \textit{GA (adaptive AB)} suggests a slight improvement over the standard GA across all architectures and initializations. However, since the p-value exceeds 0.05, this difference in performance is not statistically significant. Similarly, for Edge Popup with the varied pruning rate, the negative coefficient indicates slightly worse performance for \textit{EP (adapted)}, which supports our previous findings. Since the performances of both \textit{GA (adaptive AB)} and \textit{EP (adapted)} are not significantly different from the reference algorithms, we focus on comparing \textit{GA} and \textit{EP (50\%)}. The comparison reveals a large negative coefficient, indicating that \textit{EP (50\%)} performs significantly worse. This difference is statistically significant with a p-value of 0. When compared to backpropagation, the GA configuration shows a moderate decrease in performance, which is also statistically significant. For the \texttt{circles} dataset, the analysis presents a similar pattern. However, here, \textit{GA (adaptive AB)} has a negative coefficient, supporting the (almost statistically significant) result that the base GA configuration is more suitable for this dataset. Taking all random effects into account, backpropagation clearly outperforms the GA in this case.
\\[2pt]
We conclude this section with the general observation that the genetic algorithm has a significantly better performance than edge-popup in the considered settings. Additionally, the GA algorithm demonstrates a performance that is only moderately worse than backpropagation in terms of the final accuracy on the \texttt{moons} dataset.



\subsection{Multi-Class Performance}\label{subsec:prob_multiclass}
\begin{figure*}[t]\centering
    \subfloat[Accuracy, $R = 25$\label{fig:blobs_acc}]{\includegraphics[width=0.46\textwidth]{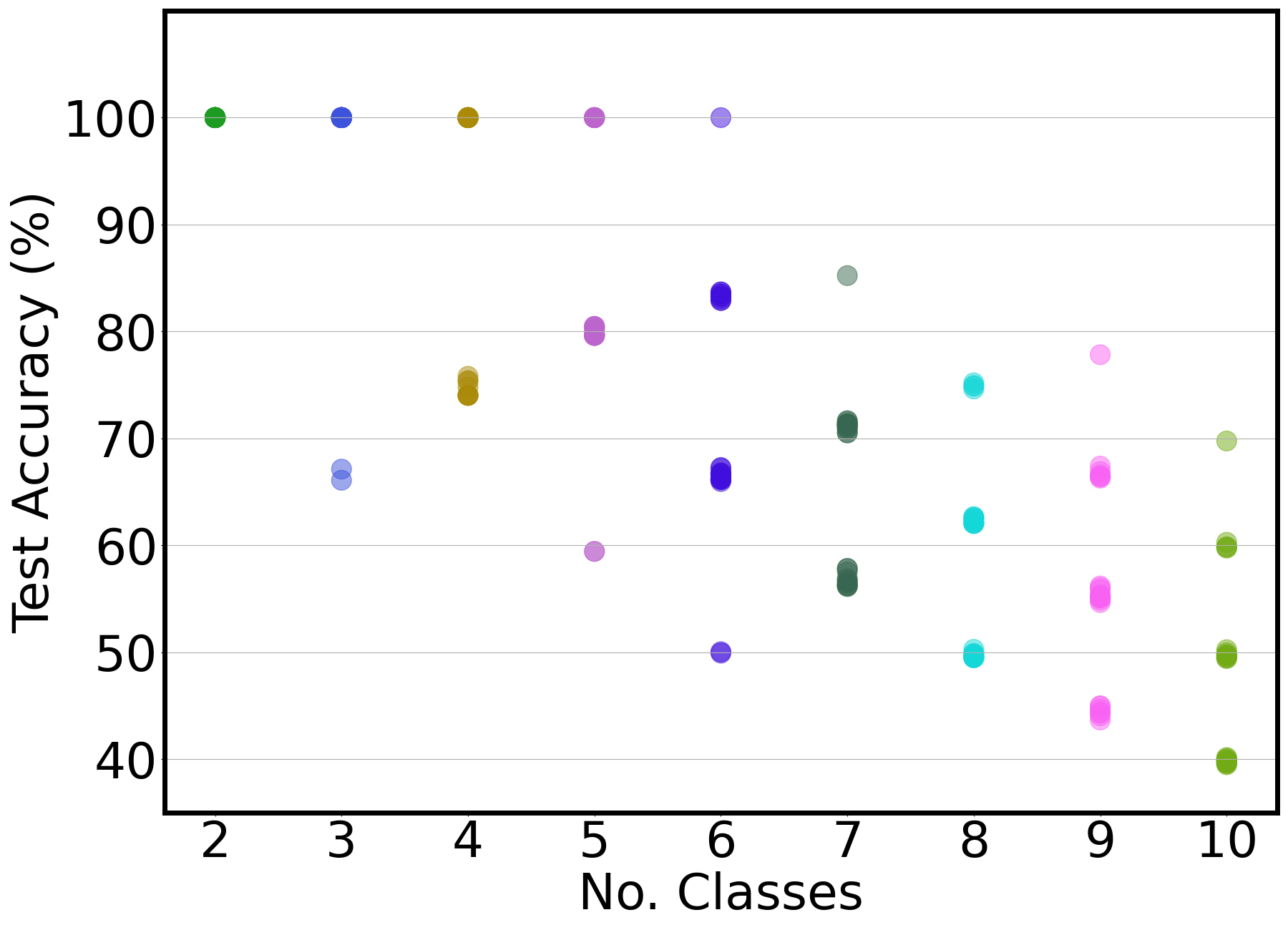}}\hspace{0.05\textwidth}
    \subfloat[Cross-Entropy Loss, $R = 25$\label{fig:blobs_loss}]{\includegraphics[width=0.46\textwidth]{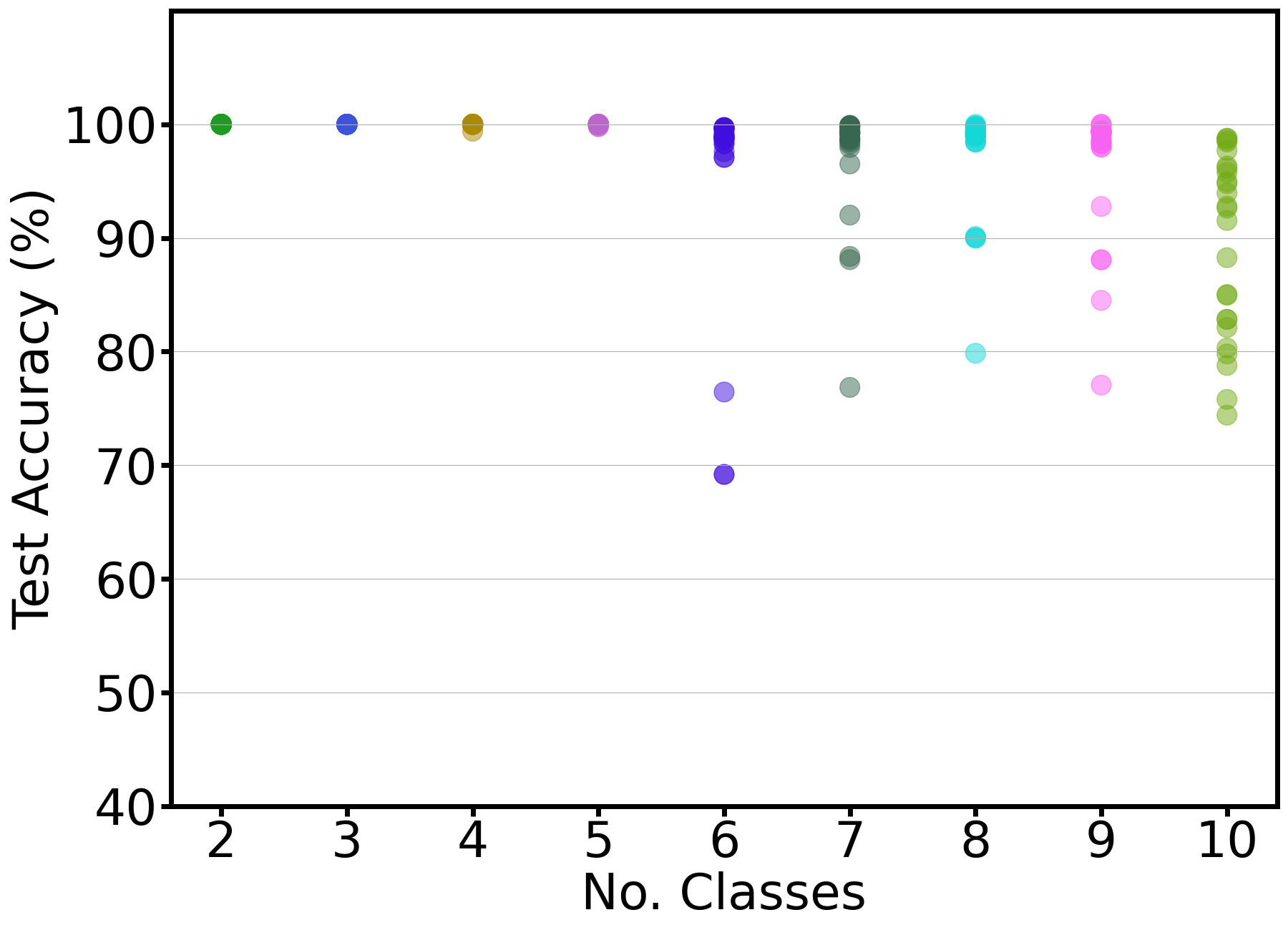}}
    \caption{Multi-Class performance on $R$ runs: (\subref{fig:blobs_acc}) Overview of the distribution of final accuracies achieved by the GA (without accuracy bound) for different variants of the \texttt{blobs} dataset when using accuracy as objective function. For the binary case we consider class labels 0 and 1, for the ternary case 0, 1, and 2, and so on. All runs were performed on architecture C (cf. Table~\ref{tab:architectures}). (\subref{fig:blobs_loss}) Performance of default GA configuration without accuracy bound on the \texttt{blobs} dataset optimizing the cross-entropy loss. Figure (\subref{fig:blobs_acc}) taken from \cite{ncta24}.}
    \label{fig:blobs_acc_loss}
\end{figure*}

So far, we only considered datasets for binary classification using accuracy as performance metric. It turns out this approach has a much harder time finding suitable lottery tickets for multi-class classification problems. There are several reasons for that. A major problem arises from the fact that the apparent correlation between a reduction in loss and an increase in accuracy, which is often observed in standard neural network training, does not generally apply in reverse. It is possible to obtain high accuracy networks that still have a high loss. A common loss function for classification tasks is the cross-entropy loss, which quantifies the model's uncertainty in assigning a sample to a specific class. Let the model's output for a sample $x_{i}$ is denoted as $\hat{y}_{i} \in [0, 1]$, which corresponds to the predicted probability of $y_{i}$ being the correct class label for $x_{i}$. In the best case, $\hat{y}_{i}$ is close to $1$ and $y_{i}$ is the correct label, resulting in a small loss. A more problematic case occurs when the model still makes the correct prediction, but its predicted probability is close to random sampling (e.g. 0.5 in the binary case). We observed that a lot of SLTs found by the GA fall into this second category, i.e. they achieve high accuracies but are very uncertain about their prediction. The uncertainty increases as the number of classes increases. The consequences of this effect on the \texttt{blobs} dataset become apparent in Fig.~\ref{fig:blobs_acc}. While for 2--5 class instances, the GA can retrieve subnetworks with perfect accuracy, trying to distinguish more class labels leads to increasingly bad final accuracies. Fortunately, this effect can be mitigated by optimizing the loss instead of the accuracy. Fig.~\ref{fig:blobs_loss} illustrates this change. We observe that the accuracies of the previously problematic runs for 7--10 clusters are now elevated and, apart from the case with 10 classes, each setting includes runs that achieve approximately 100\% test accuracy. A reason for the slight drop for the 10 classes case could be the immediate  proximity of the 9th and 10th cluster (the rightmost pink and olive colored clusters in Fig.~\ref{fig:blobs}), making their distinction considerably harder.
\\[2pt]

\noindent To better understand the difference in complexity that arises from optimizing different performance metrics, we consider the shape of the respective objective functions. A direct visualization of the optimization landscape of the GA is challenging because masking is a discrete operation, and genetic operations are non-continuous. Instead, we approximate its complexity by drawing the area of the continuous parameter space around a specific point (i.e. the parameters of the found SLT). For that, we adopt the visualization technique used in \cite{li2018visualizing} to obtain two-dimensional representations of the surfaces of the high-dimensional cross-entropy loss function and the accuracy function. Let $\mathbf{w}^s$ denote the final parameter vector of a strong lottery ticket network discovered using our genetic algorithm. Additionally, let $d_1$ and $d_2$ be two direction vectors, whose values are sampled from a Gaussian distribution with the same dimensionality as $\mathbf{w}^s$. Using two scaling factors $\delta, \eta \in \mathbb{R}$ we can shift the original network to any point in the slice of the parameter space spanned by the transformation $\mathbf{w}^t = f(\mathbf{w}^s, d_1, d_2, \delta, \eta) = \mathbf{w}^s + \delta d_1 + \eta d_2$. For the transformed network, we can calculate the loss $\mathcal{L}(\mathbf{w}^t)$ and the accuracy $\mathcal{A}(\mathbf{w}^t)$. By selecting evenly spaced values for $\delta$ and $\eta$, we can plot the SLT's loss and accuracy landscapes.

\begin{figure*}[t]\centering
    \subfloat[Loss Landscape\label{fig:blobs_loss_land}]{\includegraphics[width=0.48\textwidth]{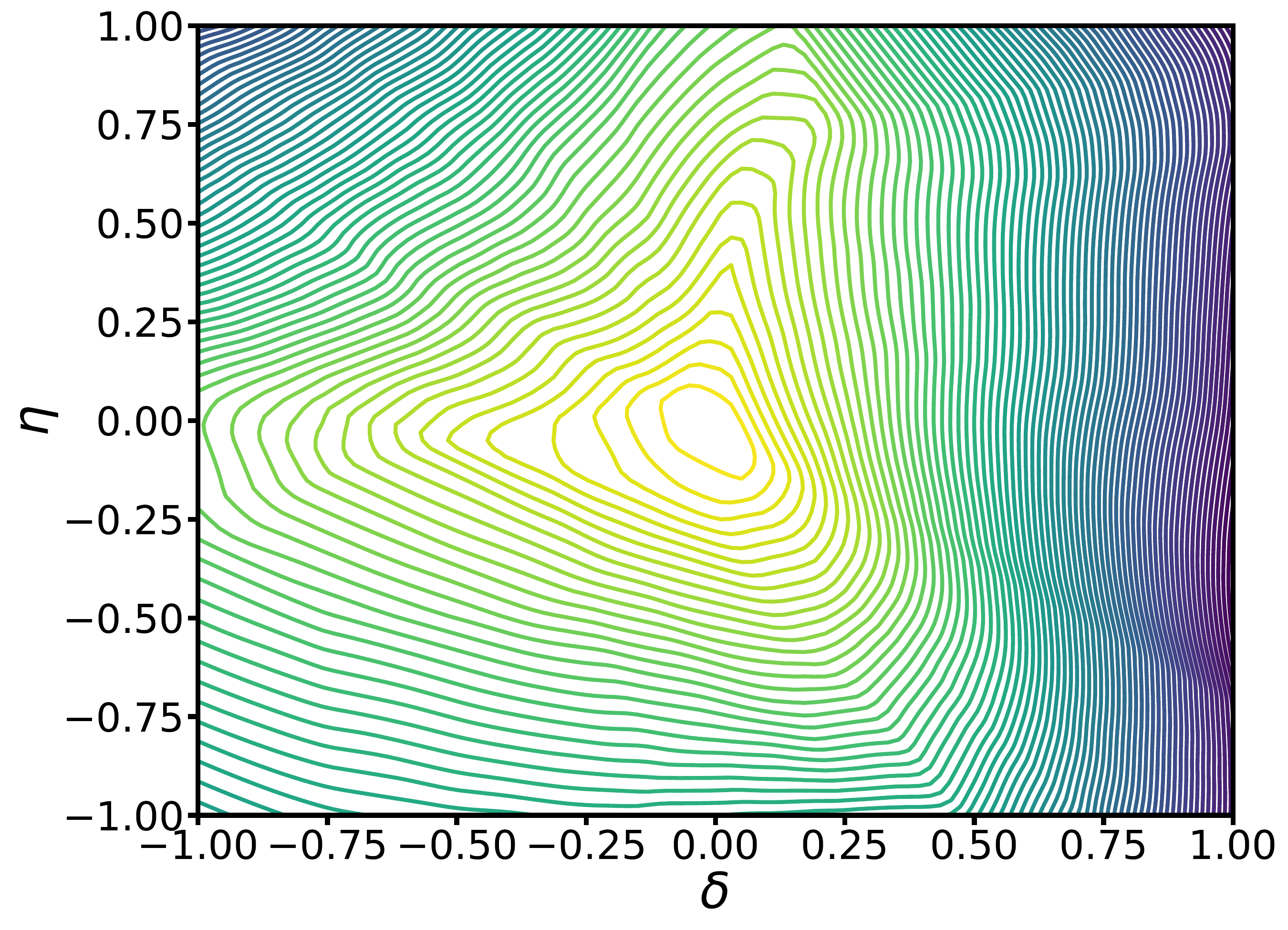}}\hspace{0.02\textwidth}
    \subfloat[Accuracy Landscape\label{fig:blobs_acc_land}]{\includegraphics[width=0.48\textwidth]{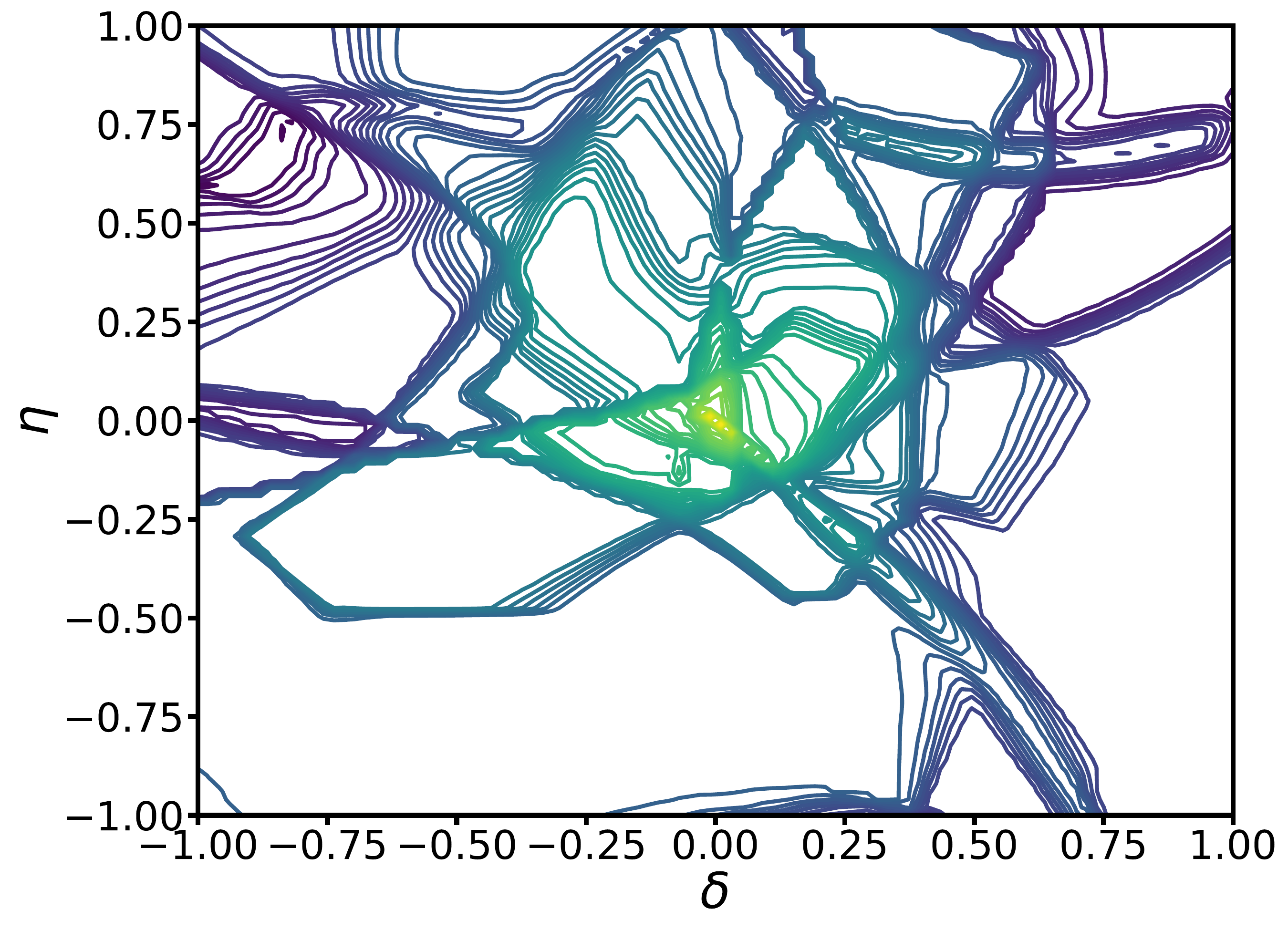}}
    \caption{Visualization of the loss landscape (\subref{fig:blobs_loss_land}) and the accuracy landscape (\subref{fig:blobs_acc_land}) of a selected run from the multi-class classification experiment on the \texttt{blobs} dataset with 10 clusters (cf. Fig.~\ref{fig:blobs_acc_loss}). $\delta$ and $\eta$ are our two scaling factors for the parameter transformation. These plots only show a small portion of the landscapes. The color gradient from purple to orange indicates an improvement of the respective objective value.}
    \label{fig:blobs_landscapes}
\end{figure*}

\noindent Fig.~\ref{fig:blobs_landscapes} shows the results for one subnetwork obtained by executing the GA on the \texttt{blobs} dataset with 10 clusters. Fig.~\ref{fig:blobs_loss_land} visualizes the loss landscape for different $\delta$ and $\eta$ values from the range $[-1,1]$. Similar to contours in maps, the lines in this plot represent different loss levels. The closer the lines are to one another, the steeper the ascent/descent. Regions with more distant contours correspond to relatively flat areas or plateaus. Darker lines represent larger loss values, brighter lines smaller ones. The original subnetwork is located in the middle of the loss landscape at point $(0,0)$. It can be seen clearly, that the SLT is located in a local minimum, with no other region exhibiting a smaller loss value. It is also apparent that the steepest ascent happens when $\delta$ is close to $1.0$. Overall the displayed region seems to be relatively convex, promising an easier optimization, at least for this slice of the parameter space. This is not the case for the accuracy landscape depicted in Fig.~\ref{fig:blobs_acc_land}. Here, brighter colors represent higher accuracies. Compared to the loss landscape, the accuracy landscape is extremely non-convex. The region with the highest accuracy is very small and is concentrated on the closest neighbors of the SLT. This implies that small changes to the parameterization of the SLT already lead to a significant drop in accuracy. The large white space areas represent plateaus with only small variability. All in all, the accuracy landscape appears to be very hostile, at least for optimization algorithms that depend on local curvature information. From this analysis, no final conclusion about the exact nature of the optimization landscapes can be drawn, but it provides a general hint about the varying complexities the GA might encounter.
\\[2pt]
\begin{figure}[t]
  \subfloat[\texttt{blobs}, $R = 25$\label{fig:blobs_comp}]{\includegraphics[width=0.42\linewidth]{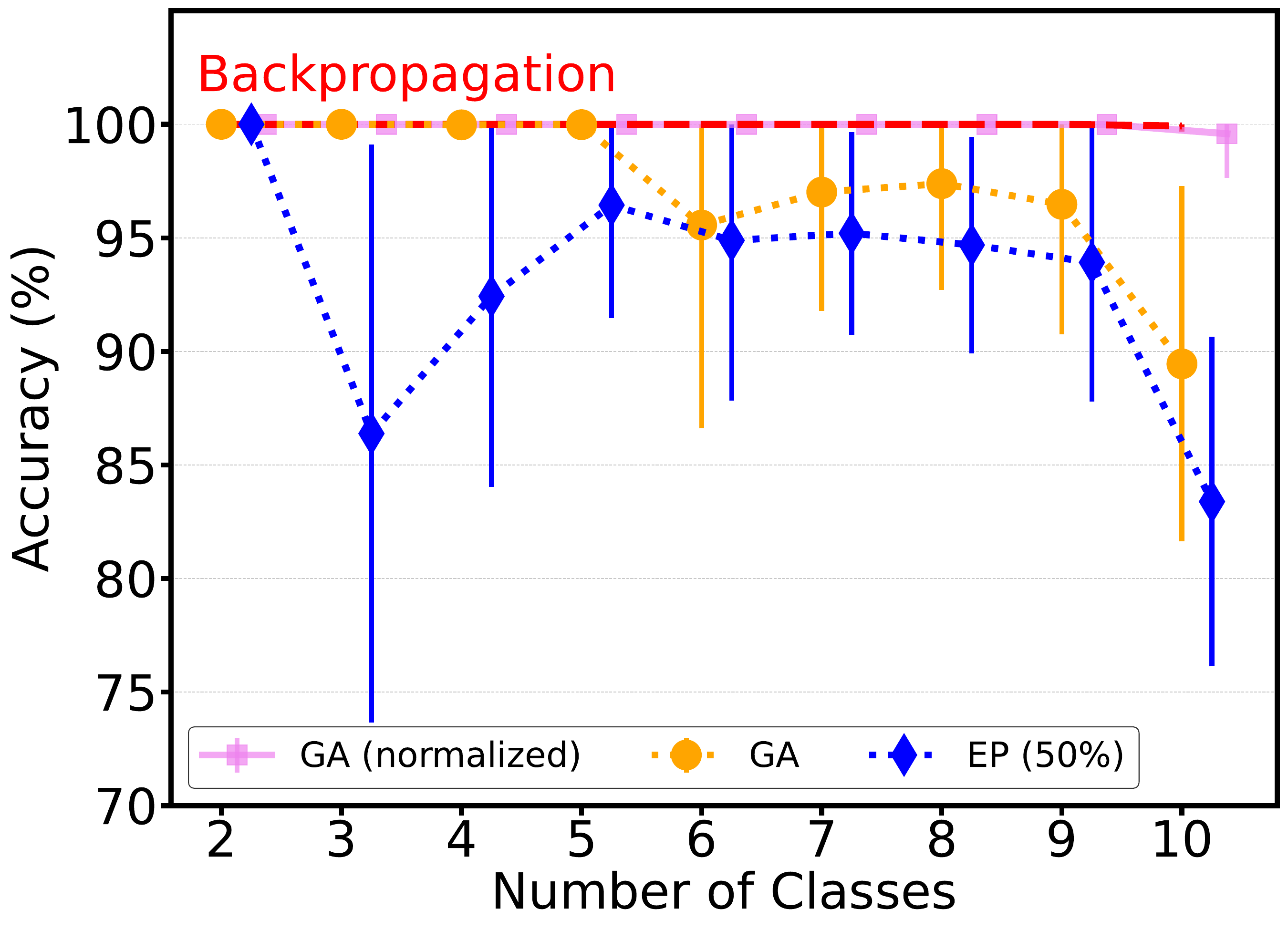}}\hspace{0.02\textwidth}
  \subfloat[Reached mean accuracies ± standard deviation\label{tab:blobs_statistics}]{\centering
  \resizebox{0.55\linewidth}{!}{
    \small\begin{tabular}{cccccc}
        \toprule
        & No. Classes & GA & EP (50\%) & Backprop. & GA (normalized) \\
        \midrule
        \parbox[t]{2mm}{\multirow{9}{*}{\rotatebox[origin=c]{90}{\texttt{blobs}}}}
        & 2 & 100.0\% ± 0.0 & 100.0\% ± 0.0 & 100.0\% ± 0.0 & 100.0\% ± 0.0 \\
        & 3 & 100.0\% ± 0.0 & 86.4\% ± 12.9 & 100.0\% ± 0.0 & 100.0\% ± 0.0 \\
        & 4 & 99.97\% ± 0.2 & 92.4\% ± 8.6 & 100.0\% ± 0.0 & 100.0\% ± 0.0 \\
        & 5 & 99.99\% ± 0.04 & 96.4\% ± 5.1 & 100.0\% ± 0.0 & 100.0\% ± 0.0 \\
        & 6 & 95.5\% ± 9.2 & 94.9\% ± 7.2 & 100.0\% ± 0.0 & 100.0\% ± 0.0 \\
        & 7 & 97.0\% ± 5.3 & 95.2\% ± 4.6 & 100.0\% ± 0.0 & 100.0\% ± 0.0 \\
        & 8 & 97.4\% ± 4.8 & 94.7\% ± 4.9 & 100.0\% ± 0.0 & 100.0\% ± 0.0 \\
        & 9 & 96.5\% ± 5.7 & 93.9\% ± 6.2 & 100.0\% ± 0.0 & 100.0\% ± 0.0 \\
        & 10 & 89.6\% ± 7.8 & 83.4\% ± 7.4 & 99.9\% ± 0.3 & 99.6\% ± 2.0 \\
        \bottomrule
        \\
        \\
    \end{tabular}}}  
  \caption{(\subref{fig:blobs_comp}) Performance comparison on the \texttt{blobs} dataset, based on $R = 25$ runs from the standard GA configuration, edge-popup with 0.5 fixed prune-rate, backpropagation utilizing the hyperparameters from Table~\ref{tab:backprop_hyperparameters} and a GA evolved on the normalized version of the dataset. The markers correspond to mean accuracy values of the runs and the vertical lines are the standard deviations. Top lines surpassing 100\% accuracy are cropped. (\subref{tab:blobs_statistics}) A detailed numerical overview of the respective mean accuracies, including the standard deviation.}
  \label{fig:blobs_alg_comp}
\end{figure}

\noindent Based on our previous results, we compare the performance of the GA, when optimizing the loss, with the performances achieved by edge-popup and networks trained via backpropagation. We use the same dataset instance of the \texttt{blobs} dataset with 2--10 clusters and evaluate the algorithms on network architecture C. \footnote{It is likely, that using a larger architecture e.g. network D would result in slightly higher accuracy values, but for this experiment we focus on the relative performance and not on the potential absolute maximum.} For edge-popup, we use the default version with 0.5 as fixed prune-rate. Apart from the general GA configuration (without the adaptive accuracy bound) we include runs obtained on a normalized version of the dataset, which we will refer to as ``GA (normalized)''. Fig.~\ref{fig:blobs_alg_comp} gives an overview of the results. The different shaped markers represent the mean accuracies, while the length of the error bars indicate the standard deviation of the runs. Networks trained with backpropagation achieve approximately 100\% accuracy for all cluster numbers. The runs of the \texttt{GA} configuration (marked in orange) are the same as those depicted in Fig.~\ref{fig:blobs_acc_loss}. Interestingly, the mean accuracy for 8 clusters is higher than for 6 and 7 clusters, with a smaller standard deviation. A potential reason for this could be that the runs with 6 and 7 clusters did not fully converge yet. Another possibility could be that the relative positions of the additional clusters helped to better differentiate previously hard to distinguish clusters. In comparison, apart from the scenario with only two clusters, edge-popup performs worse across all settings. Particularly noticeable is the sudden drop for 3 clusters, with a subsequent recovery for clusters 4 and 5. There is a lot of performance variability between the runs, implying difficulties in proper convergence for different parameter initializations. Why this distinctive behavior occurred remains open for future work. For higher cluster numbers, the performance development more closely resembles that of the GA, albeit still performing worse. A direct comparison of ``GA (normalized)'' with ``GA'' and edge-popup would be unfair, but we included its data points to highlight the strong influence on the generalization performance of subnetworks found by the GA when the dataset is normalized. To explore the reason behind this, we computed the \texttt{Hessian} of the loss functions across different runs on both normalized and non-normalized datasets to assess the general curvature of the loss landscape. One observation is that compared to ``GA (normalized)'', runs on the non-normalized dataset tend to have a few very large eigenvalues, indicating that the loss landscape around the minimum is very sharp in some directions. According to theory, this could lead to a worse generalization \cite{dinh2017sharp}.

\begin{figure}[t]
  \subfloat[\texttt{digits}, $R = 25$\label{fig:digits_comp}]{\includegraphics[width=0.42\linewidth]{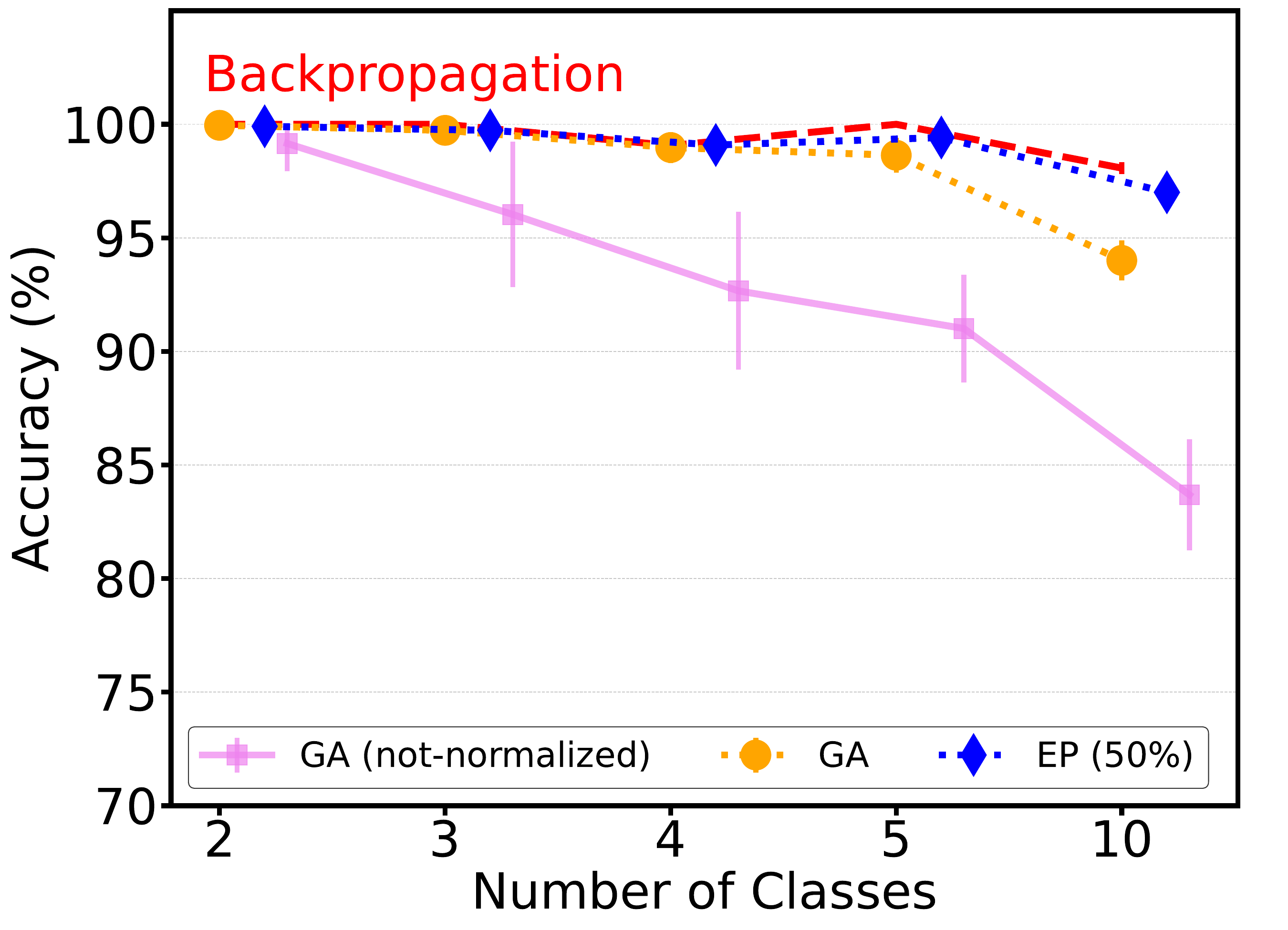}}\hspace{0.02\textwidth}
  \subfloat[Reached mean accuracies ± standard deviation\label{tab:digits_statistics}]{\centering
  \resizebox{0.55\linewidth}{!}{
    \small\begin{tabular}{cccccc}
        \toprule
        & No. Classes & GA & EP (50\%) & Backprop. & GA (not-normalized) \\
        \midrule
        \parbox[t]{2mm}{\multirow{5}{*}{\rotatebox[origin=c]{90}{\texttt{digits}}}}
         & 2 & 99.9\% ± 0.2 & 99.9\% ± 0.3 & 100.0\% ± 0.0 & 99.2\% ± 1.3 \\
         & 3 & 99.7\% ± 0.4 & 99.7\% ± 0.4 & 100.0\% ± 0.0 & 96.0\% ± 3.3 \\
         & 4 & 98.9\% ± 0.7 & 99.1\% ± 0.6 & 99.1\% ± 0.3 & 92.7\% ± 3.5 \\
         & 5 & 98.6\% ± 0.8 & 99.4\% ± 0.5 & 100.0\% ± 0.0 & 91.0\% ± 2.4 \\
         & 10 & 94.0\% ± 0.9 & 97.0\% ± 0.5 & 98.1\% ± 0.3 & 83.7\% ± 2.5 \\
        \bottomrule
        \\
        \\
    \end{tabular}}}  
  \caption{(\subref{fig:digits_comp}) Comparison of performance on the \texttt{digits} dataset, based on $R = 25$ runs using the standard GA configuration, edge-popup with a fixed prune-rate of 0.5, backpropagation with hyperparameters from Table~\ref{tab:backprop_hyperparameters}, and a GA applied to the non-normalized dataset. The markers represent the mean accuracy across the runs, while the vertical lines show the standard deviations. (\subref{tab:digits_statistics}) A detailed numerical summary of the mean accuracies and their corresponding standard deviations.}
  \label{fig:digits_alg_comp}
\end{figure}

\paragraph{Digits Dataset}
Throughout our experiments on the \texttt{blobs} dataset, we uncovered several important factors that exert a considerable influence on the efficacy of the GA when the number of clusters to differentiate increases, i.e. using a loss function as performance metric instead of directly optimizing the accuracy to reinforce the confidence in predicting class probabilities; and working on a normalized version of the dataset to reduce the risk of ``sharp minima'' which could potentially lead to better generalization. In the remainder of this section, we want to study whether the GA can find high-performing subnetworks when, in addition to a higher number of class labels, the input dimensionality is increased as well. For this purpose, we consider the \texttt{digits} dataset with up to 10 class labels ranging from $\{0, ..., 9\}$ and an input dimensionality of 64 pixel values. To adhere to the preliminary results of \cite{ncta24} we evaluate the GA and the baselines on architecture B, using 2, 3, 4, 5, and 10 class labels respectively. This results in bit masks containing $5550$ bits (for the ten-classes case) instead of the previous maximum of $400$ bits for the \texttt{blobs} dataset. We again use edge-popup with a fixed prune-rate of 0.5 and backpropagation-trained networks as baselines. In addition, we also include results for the GA on the non-normalized dataset. Fig.~\ref{fig:digits_alg_comp} depicts our findings. 
\\[2pt]
The worst performing configuration is “GA (not normalized)”, which supports our assumption regarding the importance of normalization for datasets with similar properties to those considered here. Overall, the achieved mean accuracies are considerably higher for the other algorithms compared to the \texttt{blobs} dataset. Noticeably, this experiment marks the first instance where edge-popup scored higher accuracies compared to our GA, with a maximum performance difference of around 3\% for the ten-classes case. This could indicate that our current genetic operators are prone to suboptimal local minima for high parameter counts. Determining the exact reasons and exploring the potential of more sophisticated operators remains content of future work. Finally, we observe a similar phenomenon to the one encountered for the \texttt{blobs} dataset where there is a minor decline in accuracy in the four-classes case, with higher mean accuracies achieved for the five-class setting, affecting both edge-popup and backpropagation, but not our GA.
\\[2pt]
\begin{figure*}[t]
  \centering
  \includegraphics[width=0.6\linewidth]{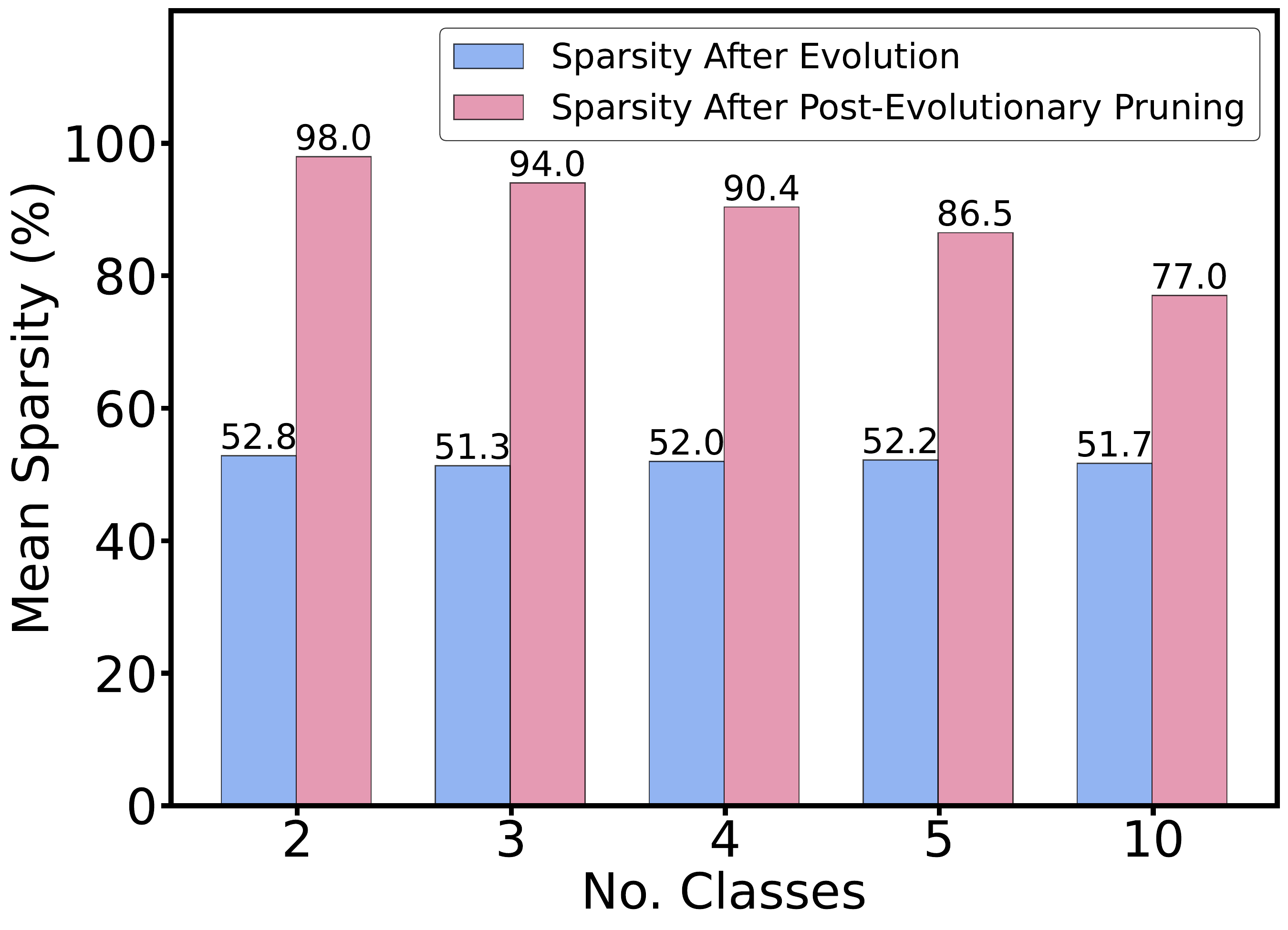}
  \caption{Illustration of the capabilities of our post-evolutionary pruning routine applied to the best performing subnetworks found by the GA on the \texttt{digits} dataset. The results are taken from the runs from the previous experiment (cf. Fig.~\ref{fig:digits_alg_comp}). Displayed are the mean sparsities of the subnetworks before and after applying the post-evolutionary pruning routine. The numbers over the bars correspond to these mean sparsity values.}
  \label{fig:sparsities_digits}
\end{figure*}

\noindent The default implementation of the generation operation of the GA (utilized in the preceding experiments) generates individuals with 50\% sparsity when constructing the initial population. For smaller network architectures, this initial sparsity is often increased significantly during the evolution (cf. architecture A, Fig.~\ref{fig:edge_moons_adapted}). The larger the architecture, the smaller this relative improvement is. For architecture D on the \texttt{moons} dataset, the increase is only about 5\%. As motivated in Section~\ref{subsec:post_evol_prune}, applying a simple post-processing routine to remove as many remaining irrelevant weights as possible after the GA has terminated can additionally reduce the number of non-zero weights in the found subnetwork. The enhanced capabilities of our pruning pipeline become evident when we consider high parameter networks like those used for the \texttt{digits} dataset. Fig.~\ref{fig:sparsities_digits} shows the sparsity development before and after applying the post-processing routine on the evolved subnetworks. The blue bars in the figure demonstrate that, in the course of evolution, the initial sparsity is increased by less than 3\% for all considered class settings. On the other hand, the potential maximum sparsity (where accuracy is not yet affected) can be much higher: Up to 98\% for the two-classes case, indicating an extreme overparameterization of the network architecture used. The achieved maximum sparsity after applying post-evolutional pruning decreases with additional class labels, but still leads to an improvement of $\approx 25.3\%$ for the ten-classes case. Given that a sequential traversal of the final pruning mask has a runtime complexity of $O(l*b^2)$, with $l*b^2$ representing the potential worst-case size of the pruning mask (cf. Section \ref{subsec:scalability}), using this additional routine does not affect the overall complexity of our algorithm, while adding substantial gains. 

\section{Conclusion}
\label{sec:conclusion}
In this work, we proposed a genetic algorithm for finding strong lottery ticket networks following a new paradigm for optimizing network parameters that leads to high accuracies without any conventional training steps. We have shown that the GA surpasses the performance of the gradient-based  state-of-the-art on three of the four considered datasets and reaches comparable accuracy levels to networks trained with backpropagation. Setting a minimum accuracy threshold via an adaptive accuracy limit when generating new individuals may lead to slight improvements, but these additional gains were found to be statistically insignificant, at least for the datasets considered. Furthermore, our experiments highlight the importance of a proper weight initialization, when using the GA and edge-popup. Although, optimizing accuracy directly did prove effective for smaller architectures on the binary classification tasks, the highly non-convex shape of the accuracy landscape observable for larger architectures with several class labels resulted in a significant drop in performance. Optimizing the cross-entropy loss instead, and normalizing the datasets, lead to final accuracies matching those of backpropagation for the \texttt{blobs} dataset even for high cluster numbers. The \texttt{digits} dataset proved to be more challenging for the GA, with slightly lower performance levels than the two baselines when the number of classes is high. Finally, we have accounted for the prioritization of the performance objective during the evolution, resulting in comparably smaller sparsity gains, by employing a ``post-evolutionary pruning'' routine that significantly increases the sparsity of the final best performing subnetwork. 
\\[2pt]
Apart from expanding our general understanding of the SLTH phenomenon and its implications, there are several opportunities for future research. Considering the importance of proper weight-initialization, it would be a promising research direction to study different parameter initializations in the context of the SLTH, both from an empirical and a theoretical viewpoint. Much of the complexity of the optimization is the result of the extremely large search space. If we could restrict our search to a promising area of the search space, e.g. by considering only a subset of the model parameters while keeping the others untouched, this would result in a much smaller number of candidate subnetworks, likely leading to faster convergence. It is to be determined if such promising areas can be identified at the start of the evolution. Additionally, it might not be necessary to consider all the training data for identifying high-performing subnetworks (cf. \citep{zhang2021efficient}). Much of the runtime in practice stems from evaluating individuals. Especially for large-scale network architectures like transformer models, this can become problematic when the dataset size is large. Apart from genetic algorithms, any method suitable for binary combinatorial optimization can in principle be applied to find strong lottery tickets. There might be other promising alternatives (cf. \citep{whitaker2022quantum}). 
\\[2pt]
It is important to note that, given the absence of gradient information in the GA, the potential applications of our approach extend beyond the realm of classical neural networks, which depend on differentiable functions. We considered accuracy and the cross-entropy loss for our performance evaluation, but using our GA it should be possible to directly incorporate non-differentiable evaluation metrics, such as \textit{edit distance} \citep{levenshtein1966binary} for string comparisons or \textit{logical consistency} checks in propositional logic — eliminating the need for potentially suboptimal differentiable surrogates (cf. \citep{patel2021feds, li2019augmenting}). This could have significant implications for areas like natural language processing and neural reasoning. Finally, we only considered simple feed-forward neural networks and relatively simple artificial classification tasks in this work. In a next step, we want to test the capabilities of the GA when applied to more complex network architectures and more demanding learning tasks. 
\\[2pt]
Our work marks an important step toward more efficient and effective network architectures by explicitly leveraging a fundamental property of overparameterized networks. We hypothesize that, beyond the existence of strong lottery tickets, other intrinsic characteristics of the parameter space can be exploited to further optimize model training. We argue that the role of subnetworks should be reconsidered as fundamental building blocks of efficient deep learning models, attributing for much of the generalization capabilities, requiring less computational resources, and enhancing scalability.

\begin{credits}
\subsubsection{\ackname} This work was partially funded by the Bavarian Ministry for Economic Affairs, Regional Development and Energy as part of a project to support the thematic development of the Institute for Cognitive Systems.

\subsubsection{\discintname}
The authors declare that there are no conflicts of interest related to this research. No financial, personal, or professional relationships have influenced the findings or conclusions presented in this work.
\end{credits}

%
%
%
%

\bibliographystyle{splncs04}
{\small
\bibliography{GALA}}

\end{document}